\documentclass[11pt]{article}
\usepackage{amsmath}

\usepackage[preprint]{acl}

\usepackage{times}
\usepackage{latexsym}
\usepackage{algorithm}
\usepackage{algpseudocode}
\usepackage[T1]{fontenc}

\usepackage[utf8]{inputenc}

\usepackage{microtype}

\usepackage{inconsolata}

\usepackage{graphicx}
\usepackage{booktabs}

\usepackage{listings}

\title{Learning to Draw ASCII Improves Spatial Reasoning in Language Models}

\author{Shiyuan Huang\thanks{Equal Contribution}, Li Liu\footnotemark[1], Jincheng He, Leilani H. Gilpin\thanks{Corresponding Author: lgilpin@ucsc.edu} \\
        University of California, Santa Cruz \\ \{shuan101, lliu112, jhe516, lgilpin\}@ucsc.edu
        }
\usepackage{xcolor}

\begin{document}
\maketitle
\begin{abstract}
When faced with complex spatial problems, humans naturally sketch layouts to organize their thinking, and the act of drawing further sharpens their understanding. In this work, we ask whether a similar principle holds for Large Language Models (LLMs): can learning to construct explicit visual layouts from spatial descriptions instill genuine spatial understanding? We introduce \textsc{Text2Space}, a dataset that pairs natural language descriptions with ground-truth ASCII grid layouts and spatial QA pairs, enabling us to separate failures in constructing spatial representations from failures in reasoning over them. We adopt ASCII because it is human-readable, operates entirely within the token space of language models, and encodes spatial relations in a structurally verifiable form. Our evaluation reveals a pronounced ``Read-Write Asymmetry'': LLMs interpret ASCII representations effectively but struggle to produce them from text, and these construction errors propagate to incorrect answers downstream. To address this limitation, we train models on layout construction (Text$\rightarrow$ASCII) and find that it significantly improves spatial reasoning from text alone, even without producing any ASCII at inference time. Combining construction with comprehension training further amplifies these gains. Crucially, these improvements transfer to three external spatial reasoning benchmarks, demonstrating that, much as sketching sharpens human spatial thinking, learning to construct explicit layouts instills spatial understanding that generalizes beyond the training format.

\end{abstract}

\section{Introduction}
\begin{figure*}
    \centering
    \includegraphics[width=\linewidth]{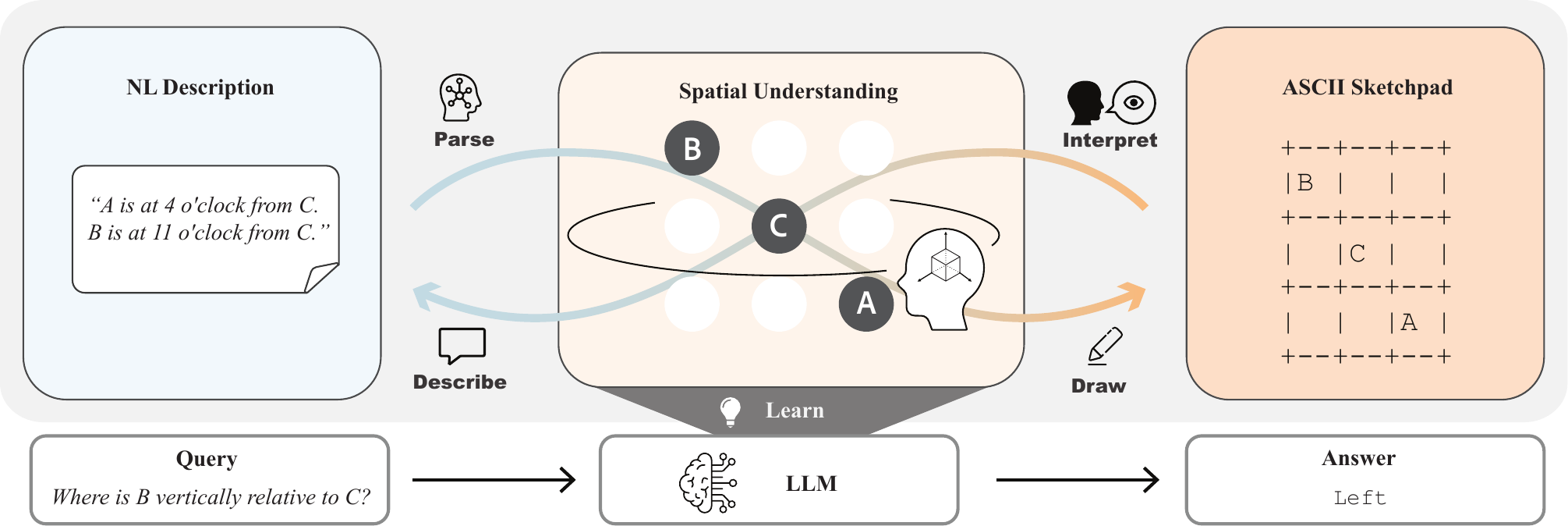}
    \caption{Overview of the study design. Inspired by empirical cognitive human strategies, we utilize Natural Language (NL) and ASCII representations as independent reasoning sources. This bidirectional cycle connects the two modalities via Parsing, Drawing, Interpreting, and Describing steps. We find that training LLMs to mimic this human-like, sketch-based reasoning process significantly improves accuracy on complex spatial reasoning tasks.}
    \label{fig:overview}
\end{figure*}
Spatial reasoning \citep{BYRNE1989564, newcombe2000making} is fundamental to human cognition \citep{malanchini2020evidence}. As large language models (LLMs) are increasingly applied to complex spatial reasoning tasks, a growing body of work has emerged to evaluate their capabilities in this domain \citep{li2024advancing, cohn2023dialectical, wang2024picture, liu2023visual, cheng2024spatialrgpt}.

However, existing benchmarks are limited to evaluating final answers \citep{yamada2023evaluating, liu2023visual}. When a model fails, it is unclear whether the error stems from a flawed spatial representation or from faulty reasoning over a correct one. More critically, even correct answers may reflect surface-level pattern matching rather than robust spatial understanding. A model can exploit lexical cues in simple configurations yet fail unpredictably as spatial complexity grows. Prior work on faithful reasoning has shown that correct answers produced through unreliable intermediate steps offer no guarantee of consistent behavior on novel problems \citep{jacovi2020towards, creswell2022faithful, parcalabescu2024measuring, xiong2025measuring}. Without a way to verify the model's intermediate spatial representation, these failure modes remain invisible.

To open this black box, we use ASCII grid layouts as an explicit, structured text that makes the reasoning process itself observable. The ASCII representation is well suited as this diagnostic lens because it lives within the model's own token space, is immediately interpretable by humans, and can be formally verified against ground truth. We introduce \textsc{Text2Space}, a benchmark of diverse spatial configurations paired with natural language descriptions, images, and three styles of ASCII grids. Each instance includes a ground-truth layout, providing verifiable supervision for both diagnosing representation failures and training models toward robust spatial understanding.
    
Figure~\ref{fig:overview} outlines our study design, which explores two processes between natural language and ASCII: construction turns text into layouts, and comprehension reads spatial relations from them. Using \textsc{Text2Space}, we contribute three investigations. First, we \textbf{benchmark} language models on ASCII comprehension, generation, and reasoning. Second, we examine the \textbf{reasoning process} by assessing whether ASCII representations faithfully support model reasoning. Most importantly, we investigate whether the spatial understanding acquired through ASCII-based learning \textbf{generalizes} to purely linguistic spatial reasoning tasks.

Our evaluation yields three principal findings that progressively identify a representation bottleneck (what we call the \emph{construction gap}) and show how to mitigate it. \textbf{Finding 1: Models can understand spatial layouts but struggle to generate them.}
Across all models, comprehending ASCII diagrams is substantially easier than constructing them from text, identifying construction as the primary constraint.
\textbf{Finding 2: This construction gap directly harms reasoning.}
When models sketch layouts as an intermediate step, their construction errors propagate to the final answer. Yet when accurate ASCII is provided instead, reasoning improves substantially over text alone, showing that the bottleneck is not reasoning itself but the quality of the representation it operates over.
\textbf{Finding 3: Training on building construction reduces the asymmetry and the benefits generalize.} Training models on layout construction improves spatial reasoning from text alone, even without producing any ASCII at inference time. Comprehension training alone does not consistently improve performance, but it leads to gains when combined with construction, producing the strongest overall results. These gains transfer to three external benchmarks, confirming that construction training builds transferable spatial understanding rather than format-specific heuristics.

We present three contributions: (1) a benchmark of spatial configurations paired with ground-truth ASCII layouts, enabling evaluation that separates representation failures from reasoning failures; (2) identification of the \emph{construction gap}, showing that models comprehend spatial layouts more reliably than they construct them and that this asymmetry propagates into downstream reasoning; and (3) empirical evidence that learning to construct spatial layouts fosters transferable internal spatial understanding, validated on three external benchmarks.

\section{Related Work}
\paragraph{Linguistic Spatial Reasoning Benchmarks.} Early work on spatial language processing focused on extracting formal representations from natural language text \citep{kordjamshidi2011spatial}. Subsequent benchmarks like bAbI \citep{weston2015towards} introduced positional reasoning, while \textsc{StepGame} \citep{stepGame2022shi} and \textsc{SpartQA} \citep{mirzaee2021spartqa} advanced the field with more complicated reasoning and broader linguistic coverage. However, a critical limitation remains across these benchmarks: they rely exclusively on final-answer accuracy. This "black-box" evaluation obscures the underlying inference process, making it impossible to distinguish between failures in representation and failures in reasoning. Our work addresses this methodological gap by introducing ground-truth intermediate representations, enabling precise diagnosis of spatial reasoning errors.
    
While integrating visual modalities (VLMs) offers a potential remedy for spatial understanding, recent studies reveal that VLMs face their own challenges in visual perception \citep{liu2023visual, liu2024right, song2024mind, kamath2023s} and spatial reasoning \citep{chen2024spatialvlm, zhang2025mitigating, ogezi2025spare}. Rather than relying on a separate visual encoder, we use ASCII grids to represent spatial layouts entirely within the language model's native token space. These text-native grids serve a dual purpose: they act not only as a reasoning source within the context but also as a verifiable representation of all spatial relations.

\paragraph{ASCII and Spatial Representation.} The utility of intermediate reasoning steps is well-established through Chain-of-Thought (CoT) prompting \citep{wei2022chain} and its extensions, such as Tree-of-Thoughts \citep{yao2023tree} and latent reasoning \citep{hao2024training}. Beyond purely linguistic traces, prior work has shown that symbolic, text-native spatial representations can effectively support spatial reasoning in language models, including grid-based worlds and discrete relational layouts \citep{cote2018textworld}. Visualization-of-Thought \citep{wu2024mind, li2025imagine} draws on human mental imagery to generate symbolic or pixel-level visualizations for spatial reasoning. While we share the intuition that explicit spatial construction aids reasoning, we focus on a different question: when and why does this process fail, and can training on it improve performance? We use ASCII as a diagnostic medium because, unlike natural language, its spatial outputs are structurally verifiable against ground truth.
    
\paragraph{Generalization of Spatial Reasoning.}
Recent research demonstrates that enhancing specific capabilities in LLMs often yields transfer benefits to other domains. For instance, \citet{shao2024deepseekmath} indicate that training on mathematics improves general reasoning benchmarks, not just quantitative tasks. Prior work shows that spatial training benefits visual reasoning tasks \citep{wei2025open, han2025learning, wang2025words, hu2025praxis}, typically through training on perception or comprehension objectives. We study a complementary mechanism: whether learning to construct spatial layouts can strengthen spatial understanding in language models, yielding transferable gains in spatial reasoning.

\begin{figure}[t]
    \centering
    \includegraphics[width=\columnwidth]{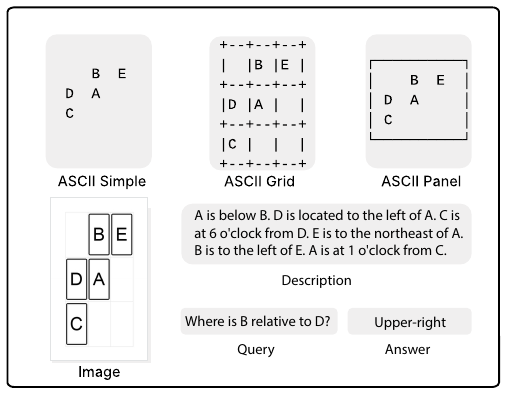} 
    \caption{An example in our \textsc{Text2Space} dataset: we have natural language description, Q-A pairs, three types of ASCII, and a rendered image.}
    \label{fig:Sample ASCII}

\end{figure}

\section{\textsc{Text2Space} Dataset}
We introduce \textsc{Text2Space}, a dataset pairing each natural language description with a ground-truth ASCII layout, a rendered image, and a query-answer pair, all derived from one spatial graph. The layout serves as a verifiable reference for construction, comprehension, and downstream reasoning. Figure~\ref{fig:Sample ASCII} shows an example.
\subsection{Data Generation}
The dataset is constructed from connected spatial graphs where entities are linked via eight directional relationships: \textit{above, below, left, right, upper-left, upper-right, lower-left, lower-right}. All relations adhere to a unified coordinate system, serving as the ground truth for both ASCII diagrams and rendered images. Instances are generated using a breadth-first search (BFS) procedure to guarantee connectivity. We enforce strict validation rules: self-loops and duplicate edges are prohibited, and any configuration conflicting with existing spatial constraints is rejected. These safeguards ensure every instance represents a valid, internally consistent topology. A detailed data generation process is presented in the Appendix~\ref{appendix:data_generation_pipeline}.

We vary complexity from simple two-component structures to intricate graphs with up to eight components and twelve relations, capturing configurations that require transitive reasoning across multiple inference steps.
Each instance includes natural language descriptions synthesized from the spatial graph, using diverse synonyms to ensure linguistic variation. Corresponding queries are designed so that target object pairs admit unique inferability, guaranteeing a single unambiguous spatial relation.

\subsection{Multi-Format Representations}
\textbf{ASCII Visualization:} We include three rendering styles to test robustness against formatting variations. The \textit{simple format} uses minimal spacing without borders for clarity. The \textit{grid format} introduces structure using \texttt{+}, \texttt{-}, and \texttt{|} characters. The \textit{panel format} employs box-drawing characters for a visually distinct layout.
\textbf{Image Representation:} The spatial arrangement is also rendered in a standardized image format with configurable grid lines and directional arrows, serving as both a multimodal evaluation target and a human-readable reference. Further dataset details and data examples are provided in Appendix~\ref{appendix:dataset_details}.

\section{Experiments}
We split \textsc{Text2Space} into training and test sets, with the test set consisting of 1,000 samples used in the following experiments. We choose the ASCII \textit{grid format} throughout our experiments; details of this format are provided in Appendix~\ref{appendix:ASCII_format}. Our experiments address three questions in sequence. \textbf{Benchmarking:} Can models both read and draw spatial layouts, and where does the gap lie? \textbf{Reasoning process:} When models sketch a layout as an intermediate step, does the quality of that sketch help or hurt downstream reasoning? \textbf{Generalization:} Does training on layout construction instill spatial understanding that transfers to text-only reasoning and external benchmarks, even when no layout is produced at inference time?
\subsection{Models}
We evaluate two model categories. \textbf{Instruct models}: including \textsc{Llama-3-70B-Instruct} \citep{llama3modelcard}, \textsc{Qwen2.5-7B-Instruct} \citep{qwen2, qwen2.5}, \textsc{Qwen3-30B-A3B-Instruct-2507}, \textsc{Qwen3-235B-A22B-Instruct-2507} \citep{qwen3technicalreport}, \textsc{GPT-4.1} \citep{openai_gpt4.1_2025}.  \textbf{Thinking model:} \textsc{gpt-oss-20B} \citep{openai2025gptoss120bgptoss20bmodel}. Unless otherwise specified, we use default settings and apply 5-shot prompting \citep{brown2020language} to ensure consistency.

\subsection{Task Setups}
Since ASCII is a unique representation that encodes two-dimensional spatial structure within a one-dimensional token sequence, we first evaluate language models’ ability to understand and generate ASCII representations. Specifically, we assess two complementary capabilities: generating an ASCII representation from a textual description (\textit{Desc$\rightarrow$ASCII}) and generating a natural-language description from an ASCII representation (\textit{ASCII$\rightarrow$Desc}).
    
We then evaluate spatial reasoning tasks, denoted $\mathcal{R}$, in which the model is given a reasoning source and asked to answer queries about spatial relationships. $\mathcal{R}_{\text{Desc}}$ refers to reasoning from textual descriptions alone, while $\mathcal{R}_{\text{ASCII}}$ refers to reasoning from ASCII representations alone.
Beyond reasoning from a single source, we investigate the role of ASCII as an intermediate step alongside descriptions through two generation orders. In $\mathcal{R}_{\text{Desc:(Ans, ASCII)}}$, the model first produces the answer and then generates an ASCII representation. In $\mathcal{R}_{\text{Desc:(ASCII, Ans)}}$, the model first generates an ASCII representation and then uses it to answer the query.
For both orders, we consider \textit{one-turn generation}, where the model produces both outputs within a single response, and \textit{two-turn generation}, where the second output is elicited via a follow-up prompt. The two generation orders mirror established self-explanation paradigms \citep{huang2023can, madsen2024self}: we compare two generation orders: predict-then-explain versus explain-then-predict \citep{camburu2018snli}.

Finally, we evaluate $\mathcal{R}_{\text{Desc+ASCII}}$. In this phase, we substitute the model-generated ASCII with the ground-truth ASCII before the model answers the query. This replacement strategy simulates an ideal reasoning process, allowing us to assess the upper-bound performance of the model when guided by an ideal spatial representation.

\subsection{Evaluation Methods}
Our evaluation protocol targets three key aspects of the model's performance:\textbf{ QA Correctness: } For the spatial reasoning QA tasks, we validate the prediction against a predefined set of relation aliases.
\textbf{Representation Alignment:} For the \textit{ASCII$\leftrightarrow$Desc} conversion tasks, we measure the correspondence between converted representations. For \textit{ASCII$\rightarrow$Desc}, the ASCII diagram is treated as the reference. We evaluate, across all generated descriptions, the percentage that can be correctly inferred from the given ASCII representation. For \textit{Desc$\rightarrow$ASCII}, we evaluate the generated ASCII diagrams, $\mathcal{R}_{\text{Desc:(Ans, ASCII)}}$ and $\mathcal{R}_{\text{Desc:(ASCII, Ans)}}$, by taking the description as the reference and measuring the proportion of the description that is correctly reflected in the ASCII output. 
\textbf{Reasoning Consistency:} For the joint generation settings, we introduce two complementary metrics to analyze internal coherence. Consistency(Cons): Measures the alignment between the generated ASCII and the generated Answer. This determines if the model’s intermediate output supports its final prediction. Consistency Accuracy(C\_Acc): Measures answer accuracy strictly on instances where the ASCII and Answer are consistent. By controlling for internal coherence, this isolates the model’s reasoning capability, helping distinguish whether errors arise from failures in constructing spatial representations or from limitations in reasoning over them. 

\subsection{Learning Spatial Relations with ASCII}
We propose that ASCII layouts function as a structured reasoning intermediate: a verifiable ``cognitive map'' that makes spatial relationships explicit. We hypothesize that learning to construct such maps \citep{lightman2023let, zhang2025lessons} consolidates spatial logic during training. If so, models trained on layout construction should improve on spatial reasoning tasks even without producing any layout at inference time, and the benefit should transfer beyond the training format.
To test and disentangle the understanding and generation of ASCII representations, we design three configurations. \textbf{Learning to Construct} (\textit{Desc$\rightarrow$ASCII}) forces the model to actively construct spatial layouts from text (performing ``mental mapping''). \textbf{Learning to Comprehend} (\textit{ASCII$\rightarrow$Desc}) focuses on parsing and verbalizing ASCII representations. \textbf{Bidirectional Alignment} (\textit{Desc$\leftrightarrow$ASCII}) combines both tasks simultaneously.

We fine-tuned \textsc{Qwen3-30B-A3B} on 4x NVIDIA A40 GPUs using the SFTTrainer from the HuggingFace TRL library~\citep{vonwerra2022trl}. We employed Low-Rank Adaptation (LoRA)~\citep{hu2021loralowrankadaptationlarge,peft} for parameter-efficient training, utilizing a split of 4,000 training and 500 validation examples. We adhered to the default library configurations for all other settings; full training details are provided in Appendix~\ref{appendix:training_details}.

We evaluate performance on the Spatial Reasoning Task ($\mathcal{R}_{\text{Desc}}$), where models answer spatial relations solely from natural language descriptions without explicit ASCII input. To assess whether the benefits extend beyond our benchmark, we further evaluate on three external tasks that each target a different facet of spatial cognition: \textsc{StepGame} \citep{stepGame2022shi,rizvi2024sparc} examines compositional directional navigation, \textsc{bAbI} task 19 \citep{weston2015towards} tests route planning through connected locations, and \textsc{SpartQA} \citep{mirzaee2021spartqa} challenges models with complex spatial questions in rich natural language. None of these tasks involve ASCII, and we evaluate the model's performance with zero-shot prompting setting. If construction training instills generalizable spatial understanding, gains should appear on these tasks despite the format mismatch.
\section{Results and Discussion}
\subsection{Benchmarking}
\label{sec:benchmarking}
\paragraph{ASCII Understanding and Generation}
Table~\ref{tab:ascii-description-translation} presents the results for the \textit{Desc$\leftrightarrow$ASCII} conversion tasks. Across all models, we observe a clear and consistent asymmetry: interpretation (\textit{ASCII$\rightarrow$Desc}) significantly outperforms generation (\textit{Desc$\rightarrow$ASCII}). This pattern suggests that while models are relatively proficient at \textbf{``reading''} or identifying patterns in ASCII, they struggle significantly more with \textbf{``drawing''} it correctly. Even the reasoning model \textsc{gpt-oss-20B} fails to construct accurate layouts in 56\% of cases, despite robust interpretative success. 

The disparity spans 13 to 27 percentage points across all evaluated models (Table~\ref{tab:ascii-description-translation}), pointing to a systematic bottleneck rather than an incidental failure mode. Interpretation relies on local pattern recognition within a provided grid, whereas generation demands global constraint satisfaction, jointly positioning all entities so that every described relation is upheld.
The construction bottleneck matters not as a standalone finding but because of its downstream consequences. As we show next, this construction deficit directly propagates to reasoning failures, and mitigating it through targeted training leads to spatial understanding that transfers beyond the ASCII format (Section~\ref{sec:learning}).

\begin{table}[t]
  \centering
  \scriptsize
  \setlength{\tabcolsep}{4.5pt}
  \renewcommand{\arraystretch}{1}
  \begin{tabular}{l c c}
    \toprule
    \textbf{Model} & \textbf{\textit{Desc $\rightarrow$ ASCII}} & \textbf{\textit{ASCII $\rightarrow$ Desc}} \\
    \midrule
    \textsc{Qwen2.5-7B}    & 16.1 & 38.4 \\
    \textsc{Llama-3-70B}   & 43.9 & 57.4 \\
    \textsc{gpt-oss-20B}   & 44.0 & 58.1 \\
    \textsc{gpt-4.1}   & 53.6 & 72.5 \\
    \textsc{Qwen3-235B} & 46.0 & 59.7 \\
    \textsc{Qwen3-30B} & 29.2 & 55.8 \\
    \addlinespace[3pt]
    \textsc{Qwen3-30B FT$_{\text{D}\rightarrow\text{A}}$} & 67.9 & 53.5 \\
    \textsc{Qwen3-30B FT$_{\text{D}\leftarrow\text{A}}$} & 24.1 & 66.5 \\
    \textsc{Qwen3-30B FT$_{\text{D}\leftrightarrow\text{A}}$} & 64.6 & 68.4 \\
    \bottomrule
  \end{tabular}
  \caption{Bidirectional mapping results. Reported scores are the percentage of correctly aligned representations across all test samples for mapping textual descriptions to ASCII (\textbf{D}esc$\rightarrow$\textbf{A}SCII) and for identifying descriptions from ASCII (\textbf{A}SCII$\rightarrow$\textbf{D}esc).}
  \label{tab:ascii-description-translation}
\end{table}

\begin{table*}[t]
  \centering
  \small
  \setlength{\tabcolsep}{6pt}
  \renewcommand{\arraystretch}{0.9}
  
  \begin{tabular}{l 
                  c 
                  c 
                  cc 
                  cc 
                  c}
  \toprule
  \textbf{Model} 
  & \multicolumn{1}{c}{\textbf{$\mathcal{R}_{\text{Desc}}$}} 
  & \multicolumn{1}{c}{\textbf{$\mathcal{R}_{\text{ASCII}}$}} 
  & \multicolumn{2}{c}{\textbf{$\mathcal{R}_{\text{Desc(Ans, ASCII)}}$}} 
  & \multicolumn{2}{c}{\textbf{$\mathcal{R}_{\text{Desc(ASCII, Ans)}}$}} 
  & \multicolumn{1}{c}{\textbf{$\mathcal{R}_{\text{Desc+ASCII}}$}} \\
  \cmidrule(lr){2-2} 
  \cmidrule(lr){3-3} 
  \cmidrule(lr){4-5} 
  \cmidrule(lr){6-7} 
  \cmidrule(lr){8-8}
  
  & Ans  
  & Ans  
  & Ans  & ASCII  
  & Ans  & ASCII  
  & Ans  \\
  \midrule
  
  \multicolumn{8}{l}{\textit{One-turn generation}} \\
  \addlinespace[2pt]
  
  \textsc{Qwen2.5-7B}        & 42.0 & 39.9 & \underline{43.6} & 12.5 & 36.7 & 12.0 & \textbf{55.5} \\
  \textsc{Llama-3-70B}       & 71.7 & \underline{75.3} & 71.1 & 42.2 & 66.9 & 41.7 & \textbf{78.4} \\
  \textsc{GPT-4.1}  & 85.8     & \underline{90.1} & 86.5     & 50.6     & 83.8     & 50.5     & \textbf{92.8} \\
  \textsc{gpt-oss-20B}       & 93.3 & 92.7 & \underline{94.3} & 38.9 & 91.9 & 42.8 & \textbf{97.0}\\
  \textsc{Qwen3-235B-A22B}   & 84.8 & \underline{89.0} & 84.2 & 45.3 & 82.3 & 45.4 & \textbf{90.1} \\
  \textsc{Qwen3-30B-A3B}     & 75.0 & \underline{83.2} & 71.1 & 27.9 & 66.6 & 27.5 & \textbf{83.7} \\
  \addlinespace[3pt]
  \textsc{Qwen3-30B-A3B FT$_{\text{Desc}\rightarrow\text{ASCII}}$}       & 82.2 & 78.6 & \underline{82.5} & 64.2 & 79.1 & 64.7 & \textbf{84.9}\\
  \textsc{Qwen3-30B-A3B FT$_{\text{Desc}\leftarrow\text{ASCII}}$}       & 73.7 & \underline{87.4} & 68.0 & 22.0 & 70.2 & 20.0 & \textbf{87.2}\\
  \textsc{Qwen3-30B-A3B FT$_{\text{Desc}\leftrightarrow\text{ASCII}}$}       & 82.7 & 81.3 & 82.9 & 60.2 & \underline{83.1} & 61.0 & \textbf{86.2} \\
  
  \addlinespace[4pt]
  \midrule
  \addlinespace[2pt]
  
  \multicolumn{8}{l}{\textit{Two-turn generation}} \\
  \addlinespace[2pt]
  
  \textsc{Qwen2.5-7B}        &        &        & \underline{43.6} & 13.5 & 42.1 & 13.7 & \textbf{52.3} \\
  \textsc{Llama-3-70B}       &        &        & \underline{70.9} & 42.8 & 69.0 & 42.3 & \textbf{78.9} \\
  \textsc{GPT-4.1}       &        &        & \underline{85.9} & 50.8 & 81.0 & 50.6 & \textbf{93.0} \\
  \textsc{gpt-oss-20B}       &        &        & \underline{93.1} & 45.0 & 87.7 & 47.9 & \textbf{94.3} \\
  \textsc{Qwen3-235B-A22B}   &        &        & \underline{85.6} & 45.1 & 77.8 & 45.3 & \textbf{89.0} \\
  \textsc{Qwen3-30B-A3B}     &        &        & \underline{74.3} & 27.4 & 69.3 & 26.4 & \textbf{82.6} \\
  \addlinespace[3pt]
  \textsc{Qwen3-30B-A3B FT$_{\text{Desc}\rightarrow\text{ASCII}}$}     &        &        & 80.7 & 65.1 & \underline{81.3} & 64.7 & \textbf{83.7} \\
  \textsc{Qwen3-30B-A3B FT$_{\text{Desc}\leftarrow\text{ASCII}}$}     &        &        & \underline{70.1} & 22.1 & 65.2 & 20.2 & \textbf{87.7} \\
  \textsc{Qwen3-30B-A3B FT$_{\text{Desc}\leftrightarrow\text{ASCII}}$}     &        &        & \underline{83.1} & 61.4 & 82.7 & 61.8 & \textbf{87.5} \\
  \bottomrule
  \end{tabular}
  
  \caption{
  Spatial reasoning accuracy across reasoning sources and settings. $\mathcal{R}_{\text{Desc(Ans, ASCII)}}$ and $\mathcal{R}_{\text{Desc(ASCII, Ans)}}$ denote whether the answer precedes or follows the ASCII output. \textit{“One-turn generation”}: answer and ASCII generated jointly. \textit{“two-turn generation”} generated in separate calls, where $\mathcal{R}_{\text{Desc+ASCII}}$ provides gold-standard ASCII. Answer accuracy: label matching; ASCII accuracy: proportion of NL relations correctly depicted. For each model, we \textbf{bold the best-performing reasoning source} and \underline{underline the second best}.}
  \label{tab:model-performance}
  \end{table*}

\paragraph{Spatial Reasoning Performance}
Table~\ref{tab:model-performance} reports accuracy on spatial reasoning tasks ($\mathcal{R}$) under different settings. On the text-only baseline ($\mathcal{R}_{\text{Desc}}$), accuracy ranges from 42.0\% (\textsc{Qwen2.5-7B}) to 93.3\% (\textsc{gpt-oss-20B}). This spread confirms that our benchmark meaningfully differentiates spatial reasoning ability, yet most models have substantial room to improve. This motivates examining whether ASCII representations can deepen spatial understanding and narrow this gap.
     
 Providing ASCII as input substantially refines performance. Most models reason more accurately from ASCII alone ($\mathcal{R}_{\text{ASCII}}$) than from text descriptions ($\mathcal{R}_{\text{Desc}}$), and the combination of both ($\mathcal{R}_{\text{Desc+ASCII}}$) yields the highest accuracy. This shows that ASCII serves as a critical signal for resolving spatial ambiguities in natural language.
 
In contrast, model-generated ASCII as an intermediate step before answering does not improve reasoning. Answering first and then drawing ($\mathcal{R}_{\text{Desc(Ans, ASCII)}}$) roughly matches the text-only baseline ($\mathcal{R}_{\text{Desc}}$), while drawing first and then answering ($\mathcal{R}_{\text{Desc(ASCII, Ans)}}$) drops below it across all base models in both \textit{one-turn} and \textit{two-turn} settings. With generated ASCII accuracy at only 12-50\% across models, the sketches are not accurate enough to serve as a promising reasoning scaffold. Since provided ASCII yields the highest accuracy while model-generated ASCII degrades it, the bottleneck is sketch quality, not the format itself. This echoes prior findings that generating intermediate artifacts can reduce task accuracy \citep{camburu2018snli, huang2023can}. However, unlike unstructured self-explanations or textual CoT \citep{wei2022chain}, where intermediate errors are difficult to detect, ASCII is structurally verifiable. This allows us to directly trace the degradation to \textbf{flawed construction} rather than treating it as a black-box effect. Models can reason over spatial layouts when provided, but cannot yet produce them reliably enough to support their own reasoning.

This diagnosis points to a concrete path forward: since the deficiency lies in representation construction, the next step is to examine whether training on this capability can strengthen the model’s spatial understanding. We first analyze whether the model’s generated layouts are internally coherent with its answers, using consistency as an unsupervised signal of reasoning reliability (Section~\ref{sec:spatial-processing}). We then investigate whether training on layout construction can instill the spatial understanding needed to bridge this discrepancy (Section~\ref{sec:learning}).

\subsection{Consistency Analysis in Spatial Reasoning}
\label{sec:spatial-processing}

Table~\ref{tab:consistency} shows the alignment between ASCII representations and textual answers across both output orders and turn settings. We report three key metrics: (1) Answer Accuracy measures overall task performance; (2) Consistency quantifies the alignment between the generated ASCII and the textual answer; and (3) Consistent Accuracy evaluates correctness specifically on instances where the representation and answer correspond.

In the \textit{one-turn} generation setting, producing the ASCII representation before the answer does not improve consistency. Similar consistency across output orders suggests that the model does not clearly use the ASCII as a reasoning scaffold, with both outputs likely reflecting the same underlying representation.
In the \textit{two-turn} generation setting, we observed an improved consistency rate, where the model first generates either the answer or the ASCII, and is then prompted to produce the other. The consistency gap between $\mathcal{R}_{\text{Desc:(Ans, ASCII)}}$ and $\mathcal{R}_{\text{Desc:(ASCII, Ans)}}$ indicates that the response to the follow-up query is strongly influenced by the previously generated output. Despite this notable improvement in consistency, answer accuracy increases only marginally. This limited gain may partly reflect to the suboptimal quality of the generated ASCII representations, which correlates with lower answer accuracy (Table~\ref{tab:model-performance}).

Nevertheless, consistency remains a valuable indicator, particularly as it requires no ground-truth annotations for either layout or the answer. While prior work has applied Natural Language Inference (NLI) to assess reasoning via consistency \citep{wang2022self, lyu2023faithful, kumar-talukdar-2020-nile}, our approach benefits from the explicit verifiability of the generated artifact. Unlike unstructured text, ASCII representations enforce strict spatial constraints and encode the full spatial configuration, not just the queried pair, allowing us to objectively assess the alignment between the representation and the answer. We observe that when the model's ASCII representation and its textual answer agree, the answer is far more likely to be correct. This consistency check requires no ground-truth labels: the ASCII itself is structurally verifiable, so agreement between the two outputs serves as an unsupervised reliability signal.

Our results show \textbf{``ASCII as a Verifier’’} as a promising direction. The structural constraints imposed by ASCII suggest that cross-representation consistency may provide a more reliable signal of reasoning validity than text-only self-consistency. Future work may explore integrating such structured verification signals into reasoning pipelines, both as an error-detection mechanism and as a guiding objective for improving intermediate representation construction.

  \begin{table}[t]
    \centering
    \scriptsize
    \setlength{\tabcolsep}{6pt} 
    \renewcommand{\arraystretch}{0.9}
    
    \begin{tabular}{l@{\hspace{2pt}}ccc@{\hspace{2pt}}ccc}
    \toprule
    \textbf{Model} 
    & \multicolumn{3}{c}{$\mathcal{R}_{\text{Desc:(Ans, ASCII)}}$} 
    & \multicolumn{3}{c}{$\mathcal{R}_{\text{Desc:(ASCII, Ans)}}$} \\
    
    \cmidrule(lr){2-4}
    \cmidrule(lr){5-7}
    
    & Ans & Cons & C\_Acc
    & Ans & Cons & C\_Acc \\
    \midrule
    
    \multicolumn{7}{l}{\textit{One-turn generation}} \\
    \addlinespace[2pt]
    
    \textsc{Qwen2.5-7B}     & 43.6 & \textbf{23.0} & 65.2 & 36.7 & 20.5 & 64.4 \\
    \textsc{Llama-3-70B}    & 71.1 & 65.2 & 87.6 & 66.9 & \textbf{65.4} & 83.8 \\
    \textsc{gpt-oss-20B}    & 94.3 & \textbf{75.9} & 97.0 & 91.9 & 75.5 & 96.9  \\
    \textsc{Qwen3-235B}     & 84.2 & 67.8 & 95.0 & 82.3 & \textbf{72.5} & 89.9 \\
    \textsc{Qwen3-30B}      & 71.1 & 46.1 & 88.5 & 66.6 & \textbf{46.3} & 82.5 \\
    \addlinespace[3pt]
    \textsc{Qwen3-30B FT$_{\text{D}\rightarrow\text{A}}$}     & 82.5 & \textbf{73.2} & 97.5 & 79.1 & 71.9 & 96.5 \\
    \textsc{Qwen3-30B FT$_{\text{D}\leftarrow\text{A}}$}     & 68.0 & 39.2 & 84.9 & 70.2 & \textbf{46.4} & 81.2 \\
    \textsc{Qwen3-30B FT$_{\text{D}\leftrightarrow\text{A}}$}     & 82.9 & \textbf{73.7} & 97.4 & 83.1 & 72.9 & 96.6 \\
    
    \addlinespace[4pt]
    \midrule
    \addlinespace[2pt]
    
    \multicolumn{7}{l}{\textit{Two-turn generation}} \\
    \addlinespace[2pt]
    
    \textsc{Qwen2.5-7B}     & 43.6 & 24.4 & 69.3 & 42.1 & \textbf{31.1} & 56.9 \\
    \textsc{Llama-3-70B}    & 70.9 & 63.9 & 89.9 & 69.0 & \textbf{71.3} & 81.9 \\
    \textsc{gpt-oss-20B}    & 93.1 & 71.8 & 96.7 & 87.7 & \textbf{77.2} & 91.6 \\
    \textsc{Qwen3-235B}     & 85.6 & 70.6 & 93.2 & 77.8 & \textbf{78.4} & 84.1 \\
    \textsc{Qwen3-30B}      & 74.3 & 50.8 & 88.8 & 69.3 & \textbf{58.6} & 75.7 \\
    \addlinespace[3pt]
    \textsc{Qwen3-30B FT$_{\text{D}\rightarrow\text{A}}$}    & 80.7 & 74.1 & 98.1 & 81.3 & \textbf{74.9} & 96.7 \\
    \textsc{Qwen3-30B FT$_{\text{D}\leftarrow\text{A}}$}    & 70.1 & 42.8 & 85.7 & 65.2 & \textbf{55.7} & 69.7 \\
    \textsc{Qwen3-30B FT$_{\text{D}\leftrightarrow\text{A}}$}   & 83.1 & 73.8 & 96.9 & 82.7 & \textbf{76.1} & 94.2 \\
    
    \bottomrule
    \end{tabular}
    
    \caption{Answer accuracy (Ans), consistency between the generated ASCII and the generated answer (Cons), and accuracy on consistent instances across different settings (C\_Acc). For each model, the higher Cons value is in \textbf{bold}. 
    }
    \label{tab:consistency}
  \end{table}

\subsection{Learning Spatial Relations with ASCII}
\label{sec:learning}

Table~\ref{tab:ascii-description-translation} details the performance on conversion tasks after fine-tuning. We note that construction training is an auxiliary task: the model learns to produce ASCII layouts, not to answer spatial questions. Construction training (FT$_{\text{D}\rightarrow\text{A}}$) raises generation accuracy (\textit{Desc$\rightarrow$ASCII}) from 29.2\% to 67.9\%, a much larger gain than comprehension training achieves (\textit{Desc$\leftarrow$ASCII}: 55.8\% to 66.5\%). The disparity confirms that ASCII generation is the primary bottleneck, with far more room for improvement. Unidirectional training improves the targeted direction but slightly degrades the other, while joint training (\textit{Desc$\leftrightarrow$ASCII}) improves both tasks (64.6\% generation, 68.4\% understanding) without this trade-off.

We now evaluate whether these improvements transfer to the downstream spatial reasoning task ($\mathcal{R}_{\text{Desc}}$), which does not  generate ASCII at inference time but requires answering spatial queries from text alone, a task format entirely absent from fine-tuning. Table~\ref{tab:model-performance} shows that construction training (FT$_{\text{D}\rightarrow\text{A}}$) improves accuracy by over 7 percentage points, indicating that learning to construct spatial layouts strengthens spatial reasoning from text alone. In contrast, comprehension training (FT$_{\text{D}\leftarrow\text{A}}$) fails to improve performance and falls slightly below the baseline, suggesting that improved comprehension without a reliable spatial representation as its foundation is not just insufficient but counterproductive. Since both conditions expose the model to identical spatial configurations, the divergence isolates \textbf{construction} as the active ingredient, not mere exposure to spatial data.

Fine-tuning on layout construction substantially improves ASCII accuracy for $\mathcal{R}_{\text{Desc:(Ans, ASCII)}}$ and $\mathcal{R}_{\text{Desc:(ASCII, Ans)}}$, reversing the pattern from Section~\ref{sec:benchmarking}: accurate sketches now scaffold reasoning rather than mislead. The bottleneck was not the representation format but \textbf{sketch quality}: even after fine-tuning, ground-truth ASCII ($\mathcal{R}_{\text{Desc+ASCII}}$) still outperforms model-generated sketches, confirming that further improvements in construction will yield additional reasoning gains.

The persistent advantage of externally provided layouts, even after construction training, parallels human spatial cognition. In a preliminary human validation (Appendix~\ref{appendix:human validation}), participants achieve substantially higher accuracy with access to ASCII layouts than with text descriptions alone. Mentally reconstructing spatial structure from sequential text is inherently more demanding than reading a provided layout for both humans and models.

The consistency analysis (Table~\ref{tab:consistency}) tells a complementary story. Construction training improves consistency between the model's sketch and its answer far more than comprehension training does. The mechanism is intuitive: when the sketch is accurate, it reinforces the textual answer; when it remains flawed, better comprehension only makes the model more aware of the conflict between its noisy sketch and the description. Construction training thus both improves consistency and ensures that consistent instances are more likely to be correct.
The confusion matrices (Appendix, Figure~\ref{fig:confusion_matrix}) corroborate this: construction training reduces confusions between spatially adjacent directions, sharpening directional boundaries.

\begin{table}[t]
  \centering
  \scriptsize
  \setlength{\tabcolsep}{3pt}
  \renewcommand{\arraystretch}{1}
  \begin{tabular}{l c c c}
    \toprule
    \textbf{Model} & \textbf{\textsc{StepGame}} & \textbf{\textsc{bAbI} Task 19} & \textbf{\textsc{SpartQA}} \\
    \midrule
    \textsc{Qwen3-30B} (base) & 54.9 & 17.8 & 26.5 \\
    \midrule
    W/ Construction (FT$_{\text{D}\rightarrow\text{A}}$) & $\mathbf{+9.9}$ & $+$17.0 & $+$43.3 \\
    W/ Comprehension (FT$_{\text{D}\leftarrow\text{A}}$) & $-$6.1 & $+$10.7 & $+$44.2 \\
    W/ Bidirectional (FT$_{\text{D}\leftrightarrow\text{A}}$) & $+$8.4 & $\mathbf{+20.6}$ & $\mathbf{+49.0}$ \\
    \bottomrule
  \end{tabular}
  \caption{Accuracy on \textsc{StepGame}, \textsc{bAbI} Task 19, and \textsc{SpartQA} under a zero-shot setting without ASCII generation. \textsc{StepGame} evaluation uses the SpaRP~PS2 1k test split. All values are deltas relative to the base model, with the best result per benchmark in \textbf{bold}.}
  \label{tab:transfer-results}
\end{table}

Table~\ref{tab:transfer-results} tests whether these gains extend beyond \textsc{Text2Space} by evaluating on three external benchmarks under zero-shot prompting, without ASCII at inference time. Both configurations that include construction training (FT$_{\text{D}\rightarrow\text{A}}$ and FT$_{\text{D}\leftrightarrow\text{A}}$) improve on all three benchmarks, while comprehension training alone does not, confirming construction as the necessary ingredient for transfer. Bidirectional training further improves over construction alone on two of three benchmarks, with 3.6 percentage points on \textsc{bAbI} task 19 and 5.7 percentage points on \textsc{SpartQA}, showing that comprehension further strengthens the foundations of spatial cognition that construction provides. Comprehension training (FT$_{\text{D}\leftarrow\text{A}}$) on its own shows an inconsistent pattern: it degrades \textsc{StepGame}, which tests precise directional navigation, yet yields large gains on \textsc{SpartQA}, which emphasizes parsing complex spatial language. This aligns with each task's demands and shows that comprehension alone cannot substitute for construction.

\section{Conclusion}
We investigate whether training language models to construct explicit spatial layouts can instill genuine spatial understanding. Using ASCII grids as a diagnostic instrument that is text-native, human-readable, and structurally verifiable, we introduce \textsc{Text2Space}, a benchmark to separate construction failures from reasoning failures.

Our results reveal a clear read-write asymmetry. Models can interpret spatial layouts effectively but struggle to construct them from text, and these construction errors propagate to downstream reasoning. Consistency analysis further shows that when a model’s ASCII representation agrees with its textual answer, answer accuracy is substantially higher. This agreement serves as an unsupervised reliability signal. Training models to construct layouts addresses this bottleneck directly: spatial reasoning from text improves even without active sketching at inference time, and the benefit transfers to three external benchmarks (\textsc{StepGame}, \textsc{bAbI}, \textsc{SpartQA}) that require spatial understanding. Only configurations that include construction training generalize across all three benchmarks, with the combination of construction and comprehension yielding the strongest gains. These results indicate that the spatial understanding gained through layout construction is not format-specific but foundational, enabling models to reason more effectively across diverse task formulations.

\section*{Limitations}
ASCII grids are one instantiation of our approach, chosen for diagnostic precision and structural verifiability. This scope entails several boundaries: we do not explore alternative formats such as coordinate lists or scene graphs, nor spatial problems involving continuous geometry, pixel-level perception, or 3D environments. As a discrete abstraction, ASCII does not capture the richness of continuous spatial perception, and our synthetic benchmark may not fully represent real-world noise and ambiguity. Future work could examine whether construction training transfers to other model families and whether richer formats scale to more complex spatial environments.

Our evaluation relies on answer accuracy and consistency between generated diagrams and predictions. These metrics provide observable, verifiable signals of reasoning behavior but do not directly reveal the model's internal representations. While ASCII's structure offers a foundation for iterative verification and self-correction, this work does not explore such mechanisms.

\newpage
\bibliography{custom}

@article{li2025imagine,
  title={Imagine while reasoning in space: Multimodal visualization-of-thought},
  author={Li, Chengzu and Wu, Wenshan and Zhang, Huanyu and Xia, Yan and Mao, Shaoguang and Dong, Li and Vuli{\'c}, Ivan and Wei, Furu},
  journal={arXiv preprint arXiv:2501.07542},
  year={2025}
}

@article{wu2024mind,
  title={Mind's eye of LLMs: visualization-of-thought elicits spatial reasoning in large language models},
  author={Wu, Wenshan and Mao, Shaoguang and Zhang, Yadong and Xia, Yan and Dong, Li and Cui, Lei and Wei, Furu},
  journal={Advances in Neural Information Processing Systems},
  volume={37},
  pages={90277--90317},
  year={2024}
}

@inproceedings{cote2018textworld,
  title={Textworld: A learning environment for text-based games},
  author={Marc-Alexandre C{\^o}t{\'e} and {\'A}kos K{\'a}d{\'a}r and Xingdi Yuan and Ben Anderson Kybartas and Tavian Barnes and Emery Fine and James Moore and Matthew J. Hausknecht and Layla El Asri and Mahmoud Adada and Wendy Tay and Adam Trischler},
  booktitle={Workshop on Computer Games},
  pages={41--75},
  year={2018},
  organization={Springer}
}

@inproceedings{li2024advancing,
  title={Advancing spatial reasoning in large language models: An in-depth evaluation and enhancement using the stepgame benchmark},
  author={Li, Fangjun and Hogg, David C and Cohn, Anthony G},
  booktitle={Proceedings of the AAAI Conference on Artificial Intelligence},
  volume={38},
  number={17},
  pages={18500--18507},
  year={2024}
}

@article{zhang2025lessons,
  title={The lessons of developing process reward models in mathematical reasoning},
  author={Zhang, Zhenru and Zheng, Chujie and Wu, Yangzhen and Zhang, Beichen and Lin, Runji and Yu, Bowen and Liu, Dayiheng and Zhou, Jingren and Lin, Junyang},
  journal={arXiv preprint arXiv:2501.07301},
  year={2025}
}

@article{kordjamshidi2011spatial,
  title={Spatial role labeling: Towards extraction of spatial relations from natural language},
  author={Kordjamshidi, Parisa and Van Otterlo, Martijn and Moens, Marie-Francine},
  journal={ACM Transactions on Speech and Language Processing (TSLP)},
  volume={8},
  number={3},
  pages={1--36},
  year={2011},
  publisher={ACM New York, NY, USA}
}

@article{weston2015towards,
  title={Towards ai-complete question answering: A set of prerequisite toy tasks},
  author={Weston, Jason and Bordes, Antoine and Chopra, Sumit and Rush, Alexander M and Van Merri{\"e}nboer, Bart and Joulin, Armand and Mikolov, Tomas},
  journal={arXiv preprint arXiv:1502.05698},
  year={2015}
}

@article{mirzaee2021spartqa,
  title={Spartqa:: A textual question answering benchmark for spatial reasoning},
  author={Mirzaee, Roshanak and Faghihi, Hossein Rajaby and Ning, Qiang and Kordjmashidi, Parisa},
  journal={arXiv preprint arXiv:2104.05832},
  year={2021}
}

@inproceedings{chen2024spatialvlm,
  title={Spatialvlm: Endowing vision-language models with spatial reasoning capabilities},
  author={Chen, Boyuan and Xu, Zhuo and Kirmani, Sean and Ichter, Brain and Sadigh, Dorsa and Guibas, Leonidas and Xia, Fei},
  booktitle={Proceedings of the IEEE/CVF Conference on Computer Vision and Pattern Recognition},
  pages={14455--14465},
  year={2024}
}

@article{wei2022chain,
  title={Chain-of-thought prompting elicits reasoning in large language models},
  author={Wei, Jason and Wang, Xuezhi and Schuurmans, Dale and Bosma, Maarten and Ichter, Brian and Xia, Fei and Chi, Ed H. and Le, Quoc V. and Zhou, Denny},
  journal={Advances in neural information processing systems},
  volume={35},
  pages={24824--24837},
  year={2022}
}

@article{yao2023tree,
  title={Tree of thoughts: Deliberate problem solving with large language models},
  author={Yao, Shunyu and Yu, Dian and Zhao, Jeffrey and Shafran, Izhak and Griffiths, Tom and Cao, Yuan and Narasimhan, Karthik},
  journal={Advances in neural information processing systems},
  volume={36},
  pages={11809--11822},
  year={2023}
}

@article{hao2024training,
  title={Training large language models to reason in a continuous latent space},
  author={Hao, Shibo and Sukhbaatar, Sainbayar and Su, DiJia and Li, Xian and Hu, Zhiting and Weston, Jason and Tian, Yuandong},
  journal={arXiv preprint arXiv:2412.06769},
  year={2024}
}

@article{liu2024right,
  title={Right this way: Can VLMs Guide Us to See More to Answer Questions?},
  author={Liu, Li and Yang, Diji and Zhong, Sijia and Tholeti, Kalyana Suma Sree and Ding, Lei and Zhang, Yi and Gilpin, Leilani},
  journal={Advances in Neural Information Processing Systems},
  volume={37},
  pages={132946--132976},
  year={2024}
}

@article{zhang2025mitigating,
  title={Mitigating spatial hallucination in large language models for path planning via prompt engineering},
  author={Zhang, Hongjie and Deng, Hourui and Ou, Jie and Feng, Chaosheng},
  journal={Scientific Reports},
  volume={15},
  number={1},
  pages={8881},
  year={2025},
  publisher={Nature Publishing Group UK London}
}

@article{llama3modelcard,
    title={Llama 3 Model Card},
    author={AI@Meta},
    year={2024},
    url = {https://github.com/meta-llama/llama3/blob/main/MODEL_CARD.md}
}

@misc{qwen2.5,
    title = {Qwen2.5: A Party of Foundation Models},
    url = {https://qwenlm.github.io/blog/qwen2.5/},
    author = {Qwen Team},
    month = {September},
    year = {2024}
}

@article{qwen2,
      title={Qwen2 Technical Report}, 
      author={An Yang and Baosong Yang and Binyuan Hui and Bo Zheng and Bowen Yu and Chang Zhou and Chengpeng Li and Chengyuan Li and Dayiheng Liu and Fei Huang and Guanting Dong and Haoran Wei and Huan Lin and Jialong Tang and Jialin Wang and Jian Yang and Jianhong Tu and Jianwei Zhang and Jianxin Ma and Jin Xu and Jingren Zhou and Jinze Bai and Jinzheng He and Junyang Lin and Kai Dang and Keming Lu and Keqin Chen and Kexin Yang and Mei Li and Mingfeng Xue and Na Ni and Pei Zhang and Peng Wang and Ru Peng and Rui Men and Ruize Gao and Runji Lin and Shijie Wang and Shuai Bai and Sinan Tan and Tianhang Zhu and Tianhao Li and Tianyu Liu and Wenbin Ge and Xiaodong Deng and Xiaohuan Zhou and Xingzhang Ren and Xinyu Zhang and Xipin Wei and Xuancheng Ren and Yang Fan and Yang Yao and Yichang Zhang and Yu Wan and Yunfei Chu and Yuqiong Liu and Zeyu Cui and Zhenru Zhang and Zhihao Fan},
      journal={arXiv preprint arXiv:2407.10671},
      year={2024}
}

@misc{qwen3technicalreport,
      title={Qwen3 Technical Report}, 
      author={Qwen Team},
      year={2025},
      eprint={2505.09388},
      archivePrefix={arXiv},
      primaryClass={cs.CL},
      url={https://arxiv.org/abs/2505.09388}, 
}

@misc{openai2025gptoss120bgptoss20bmodel,
      title={gpt-oss-120b \& gpt-oss-20b Model Card}, 
      author={OpenAI},
      year={2025},
      eprint={2508.10925},
      archivePrefix={arXiv},
      primaryClass={cs.CL},
      url={https://arxiv.org/abs/2508.10925}, 
}

@article{brown2020language,
  title={Language models are few-shot learners},
  author={Tom B. Brown and Benjamin Mann and Nick Ryder and Melanie Subbiah and Jared Kaplan and Prafulla Dhariwal and Arvind Neelakantan and Pranav Shyam and Girish Sastry and Amanda Askell and Sandhini Agarwal and Ariel Herbert-Voss and Gretchen Krueger and T. J. Henighan and Rewon Child and Aditya Ramesh and Daniel M. Ziegler and Jeff Wu and Clemens Winter and Christopher Hesse and Mark Chen and Eric Sigler and Ma-teusz Litwin and Scott Gray and Benjamin Chess and Jack Clark and Christopher Berner and Sam McCandlish and Alec Radford and Ilya Sutskever and Dario Amodei},
  journal={Advances in neural information processing systems},
  volume={33},
  pages={1877--1901},
  year={2020}
}

@article{BYRNE1989564,
title = {Spatial reasoning},
journal = {Journal of Memory and Language},
volume = {28},
number = {5},
pages = {564-575},
year = {1989},
issn = {0749-596X},
doi = {https://doi.org/10.1016/0749-596X(89)90013-2},
url = {https://www.sciencedirect.com/science/article/pii/0749596X89900132},
author = {Ruth M.J Byrne and P.N Johnson-Laird}
}

@article{yamada2023evaluating,
  title={Evaluating spatial understanding of large language models},
  author={Yamada, Yutaro and Bao, Yihan and Lampinen, Andrew K and Kasai, Jungo and Yildirim, Ilker},
  journal={arXiv preprint arXiv:2310.14540},
  year={2023}
}

@inproceedings{parcalabescu2024measuring,
  title={On measuring faithfulness or self-consistency of natural language explanations},
  author={Parcalabescu, Letitia and Frank, Anette},
  booktitle={Proceedings of the 62nd Annual Meeting of the Association for Computational Linguistics (Volume 1: Long Papers)},
  pages={6048--6089},
  year={2024}
}

@article{jacovi2020towards,
  title={Towards faithfully interpretable NLP systems: How should we define and evaluate faithfulness?},
  author={Jacovi, Alon and Goldberg, Yoav},
  journal={arXiv preprint arXiv:2004.03685},
  year={2020}
}

@article{creswell2022faithful,
  title={Faithful reasoning using large language models},
  author={Creswell, Antonia and Shanahan, Murray},
  journal={arXiv preprint arXiv:2208.14271},
  year={2022}
}

@inproceedings{lightman2023let,
  title={Let's verify step by step},
  author={Lightman, Hunter and Kosaraju, Vineet and Burda, Yuri and Edwards, Harrison and Baker, Bowen and Lee, Teddy and Leike, Jan and Schulman, John and Sutskever, Ilya and Cobbe, Karl},
  booktitle={The Twelfth International Conference on Learning Representations},
  year={2023}
}

@book{newcombe2000making,
  title={Making space: The development of spatial representation and reasoning},
  author={Newcombe, Nora and Huttenlocher, Janellen},
  year={2000},
  publisher={MIT press}
}

@article{cohn2023dialectical,
  title={Dialectical language model evaluation: An initial appraisal of the commonsense spatial reasoning abilities of LLMs},
  author={Cohn, Anthony G and Hernandez-Orallo, Jose},
  journal={arXiv preprint arXiv:2304.11164},
  year={2023}
}

@article{wang2024picture,
  title={Is a picture worth a thousand words? delving into spatial reasoning for vision language models},
  author={Wang, Jiayu and Ming, Yifei and Shi, Zhenmei and Vineet, Vibhav and Wang, Xin and Li, Sharon and Joshi, Neel},
  journal={Advances in Neural Information Processing Systems},
  volume={37},
  pages={75392--75421},
  year={2024}
}

@article{cheng2024spatialrgpt,
  title={Spatialrgpt: Grounded spatial reasoning in vision-language models},
  author={Cheng, An-Chieh and Yin, Hongxu and Fu, Yang and Guo, Qiushan and Yang, Ruihan and Kautz, Jan and Wang, Xiaolong and Liu, Sifei},
  journal={Advances in Neural Information Processing Systems},
  volume={37},
  pages={135062--135093},
  year={2024}
}

@article{liu2023visual,
  title={Visual spatial reasoning},
  author={Liu, Fangyu and Emerson, Guy and Collier, Nigel},
  journal={Transactions of the Association for Computational Linguistics},
  volume={11},
  pages={635--651},
  year={2023},
  publisher={MIT Press One Broadway, 12th Floor, Cambridge, Massachusetts 02142, USA~…}
}

@article{song2024mind,
  title={Mind the gap: Examining the self-improvement capabilities of large language models},
  author={Song, Yuda and Zhang, Hanlin and Eisenach, Carson and Kakade, Sham and Foster, Dean and Ghai, Udaya},
  journal={arXiv preprint arXiv:2412.02674},
  year={2024}
}

@inproceedings{stepGame2022shi,
title={StepGame: A New Benchmark for Robust Multi-Hop Spatial Reasoning in Texts},
author={Shi, Zhengxiang and Zhang, Qiang and Lipani, Aldo},
volume={36},
url={https://ojs.aaai.org/index.php/AAAI/article/view/21383},
DOI={10.1609/aaai.v36i10.21383}, 
booktitle={Proceedings of the AAAI Conference on Artificial Intelligence},
year={2022},
month={Jun.},
pages={11321-11329}
}

@article{camburu2018snli,
  title={e-snli: Natural language inference with natural language explanations},
  author={Camburu, Oana-Maria and Rockt{\"a}schel, Tim and Lukasiewicz, Thomas and Blunsom, Phil},
  journal={Advances in Neural Information Processing Systems},
  volume={31},
  year={2018}
}

@article{huang2023can,
  title={Can large language models explain themselves? a study of llm-generated self-explanations},
  author={Huang, Shiyuan and Mamidanna, Siddarth and Jangam, Shreedhar and Zhou, Yilun and Gilpin, Leilani H},
  journal={arXiv preprint arXiv:2310.11207},
  year={2023}
}

@article{ogezi2025spare,
  title={SpaRE: Enhancing Spatial Reasoning in Vision-Language Models with Synthetic Data},
  author={Ogezi, Michael and Shi, Freda},
  journal={arXiv preprint arXiv:2504.20648},
  year={2025}
}

@article{kamath2023s,
  title={What's" up" with vision-language models? investigating their struggle with spatial reasoning},
  author={Kamath, Amita and Hessel, Jack and Chang, Kai-Wei},
  journal={arXiv preprint arXiv:2310.19785},
  year={2023}
}

@misc{openai_gpt4.1_2025,
  author       = {{OpenAI}},
  title        = {GPT-4.1 Model Documentation},
  year         = {2025},
  howpublished = {\url{https://platform.openai.com/docs/models/gpt-4.1}},
  note         = {Accessed: 2025-12-20}
}

@article{shao2024deepseekmath,
  title={DeepSeekMath: Pushing the Limits of Mathematical Reasoning in Open Language Models},
  author={Zhihong Shao and Peiyi Wang and Qihao Zhu and Runxin Xu and Jun-Mei Song and Mingchuan Zhang and Y. K. Li and Yu Wu and Daya Guo},
  journal={ArXiv},
  year={2024},
  volume={abs/2402.03300},
  url={https://api.semanticscholar.org/CorpusID:267412607}
}

@article{wei2025open,
  title={Open vision reasoner: Transferring linguistic cognitive behavior for visual reasoning},
author={Yana Wei and Liang Zhao and Jian‐Yuan Sun and Kangheng Lin and Jisheng Yin and Jingcheng Hu and Yinmin Zhang and En Yu and Haoran Lv and Zejia Weng and Jia Wang and Chunrui Han and Yuang Peng and Qi Han and Zheng Ge and Xiangyu Tony Zhang and Daxin Jiang and Vishal M. Patel},
  journal={arXiv preprint arXiv:2507.05255},
  year={2025}
}

@article{han2025learning,
  title={Learning to see before seeing: Demystifying llm visual priors from language pre-training},
  author={Han, Junlin and Tong, Shengbang and Fan, David and Ren, Yufan and Sinha, Koustuv and Torr, Philip and Kokkinos, Filippos},
  journal={arXiv preprint arXiv:2509.26625},
  year={2025}
}

@article{wang2025words,
  title={Words That Make Language Models Perceive},
  author={Wang, Sophie L and Isola, Phillip and Cheung, Brian},
  journal={arXiv preprint arXiv:2510.02425},
  year={2025}
}

@article{hu2025praxis,
  title={Praxis-VLM: Vision-Grounded Decision Making via Text-Driven Reinforcement Learning},
  author={Hu, Zhe and Li, Jing and Pu, Zhongzhu and Chan, Hou Pong and Yin, Yu},
  journal={arXiv preprint arXiv:2503.16965},
  year={2025}
}

@article{wang2022self,
  title={Self-consistency improves chain of thought reasoning in language models},
  author={Wang, Xuezhi and Wei, Jason and Schuurmans, Dale and Le, Quoc and Chi, Ed and Narang, Sharan and Chowdhery, Aakanksha and Zhou, Denny},
  journal={arXiv preprint arXiv:2203.11171},
  year={2022}
}

@inproceedings{lyu2023faithful,
  title={Faithful chain-of-thought reasoning},
  author={Lyu, Qing and Havaldar, Shreya and Stein, Adam and Zhang, Li and Rao, Delip and Wong, Eric and Apidianaki, Marianna and Callison-Burch, Chris},
  booktitle={The 13th International Joint Conference on Natural Language Processing and the 3rd Conference of the Asia-Pacific Chapter of the Association for Computational Linguistics (IJCNLP-AACL 2023)},
  year={2023}
}

@inproceedings{kumar-talukdar-2020-nile,
    title = "{NILE} : Natural Language Inference with Faithful Natural Language Explanations",
    author = "Kumar, Sawan  and
      Talukdar, Partha",
    editor = "Jurafsky, Dan  and
      Chai, Joyce  and
      Schluter, Natalie  and
      Tetreault, Joel",
    booktitle = "Proceedings of the 58th Annual Meeting of the Association for Computational Linguistics",
    month = jul,
    year = "2020",
    address = "Online",
    publisher = "Association for Computational Linguistics",
    url = "https://aclanthology.org/2020.acl-main.771/",
    doi = "10.18653/v1/2020.acl-main.771",
    pages = "8730--8742",
}

@misc{hu2021loralowrankadaptationlarge,
      title={LoRA: Low-Rank Adaptation of Large Language Models}, 
      author={Edward J. Hu and Yelong Shen and Phillip Wallis and Zeyuan Allen-Zhu and Yuanzhi Li and Shean Wang and Lu Wang and Weizhu Chen},
      year={2021},
      eprint={2106.09685},
      archivePrefix={arXiv},
      primaryClass={cs.CL},
      url={https://arxiv.org/abs/2106.09685}, 
}

@misc{vonwerra2022trl,
  author = {Leandro von Werra and Younes Belkada and Lewis Tunstall and Edward Beeching and Tristan Thrush and Nathan Lambert and Shengyi Huang and Kashif Rasul and Quentin Gallouédec},
  title = {TRL: Transformer Reinforcement Learning},
  year = {2020},
  publisher = {GitHub},
  journal = {GitHub repository},
  howpublished = {\url{https://github.com/huggingface/trl}}
}

@Misc{peft,
  title =        {{PEFT}: State-of-the-art Parameter-Efficient Fine-Tuning methods},
  author =       {Sourab Mangrulkar and Sylvain Gugger and Lysandre Debut and Younes Belkada and Sayak Paul and Benjamin Bossan and Marian Tietz},
  howpublished = {\url{https://github.com/huggingface/peft}},
  year =         {2022}
}

@article{malanchini2020evidence,
  title={Evidence for a unitary structure of spatial cognition beyond general intelligence},
  author={Malanchini, Margherita and Rimfeld, Kaili and Shakeshaft, Nicholas G and McMillan, Andrew and Schofield, Kerry L and Rodic, Maja and Rossi, Valerio and Kovas, Yulia and Dale, Philip S and Tucker-Drob, Elliot M and Plomin, Robert},
  journal={npj Science of Learning},
  volume={5},
  number={1},
  pages={9},
  year={2020},
  publisher={Nature Publishing Group UK London}
}

@article{madsen2024self,
  title={Are self-explanations from Large Language Models faithful?},
  author={Madsen, Andreas and Chandar, Sarath and Reddy, Siva},
  journal={arXiv preprint arXiv:2401.07927},
  year={2024}
}

@article{xiong2025measuring,
  title={Measuring the faithfulness of thinking drafts in large reasoning models},
  author={Xiong, Zidi and Chen, Shan and Qi, Zhenting and Lakkaraju, Himabindu},
  journal={arXiv preprint arXiv:2505.13774},
  year={2025}
}

@inproceedings{rizvi2024sparc,
  title={SpaRC and SpaRP: Spatial reasoning characterization and path generation for understanding spatial reasoning capability of large language models},
  author={Rizvi, Md Imbesat and Zhu, Xiaodan and Gurevych, Iryna},
  booktitle={Proceedings of the 62nd Annual Meeting of the Association for Computational Linguistics (Volume 1: Long Papers)},
  pages={4750--4767},
  year={2024}
}

\newpage
\appendix
\lstset{
  basicstyle=\ttfamily\small, 
  breaklines=true,
  breakatwhitespace=false,
  aboveskip=0pt,
  belowskip=0pt,
  lineskip=-1pt,
  xleftmargin=0pt,
  xrightmargin=0pt,
  breakindent=0pt,
}

\clearpage
\section*{Appendix}

\section{Dataset Details}
\label{appendix:dataset_details}

\begin{figure*}
    \centering
    \includegraphics[width=\linewidth]{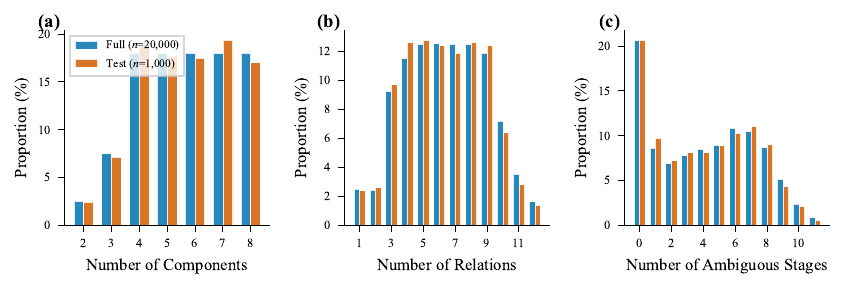}
    \caption{Distribution of (a) number of components, (b) number of relations, and (c) number of ambiguous stages. An ambiguous stage occurs when a newly introduced relation does not uniquely determine the position of an entity relative to those already placed; higher counts indicate configurations requiring more transitive inference steps. The dataset contains 20,000 instances with an average of 5.7 components and 6.4 relations.}
    \label{fig:dataset_stats}
\end{figure*}

Table~\ref{tab:dataset schema} shows possible values for each field in \textsc{Text2Space}. The dataset is in English, covering various terminologies including cardinal directions (e.g., `North'), clock-face references (e.g., `3 o'clock'), and standard spatial prepositions (e.g., `left of'). Figure~\ref{fig:dataset_stats} and Table ~\ref{Dataset_stats_in_table} provides data statistics for the full dataset and test set.

\begin{table}[h]
\centering
\begin{tabular}{ll}
\toprule
\textbf{Field} & \textbf{Possible Values} \\
\midrule
query\_type & full, vertical, horizontal \\
label (full) & above, below, left, right, \\
& upper-left, upper-right, \\
& lower-left, lower-right \\
label (vertical) & above, below, same level \\
label (horizontal) & left, right, same column \\
terminology\_used & spatial, cardinal, clock,\\
& spatial+cardinal, \\
& spatial+clock,\\
& cardinal+clock, \\
& spatial+cardinal+clock \\
ascii & simple, grid, panel \\
num\_components & 2--8 \\
num\_relations & 1--12 \\
is\_directly\_stated & true, false \\
has\_unique\_layout & true, false \\
ambiguous\_stages & 0--11\\
\bottomrule
\end{tabular}
\caption{\textsc{Text2Space} Dataset Schema, including spatial query types, terminology variations, and structural constraints (components and relations).}
\label{tab:dataset schema}
\end{table}

\section{ASCII Format Choice}
\label{appendix:ASCII_format}
Using three different ASCII representations as input, we conduct two experiments to determine the most suitable format. The first experiment evaluates the interpretation of natural language descriptions from ASCII (\textit{ASCII $\rightarrow$ Desc}), while the second assesses spatial reasoning directly from ASCII ($\mathcal{R}_{\text{ASCII}}$). As shown in Table~\ref{tab:ascii-format-result}, performance on \textit{ASCII $\rightarrow$ Desc} is comparable across all three formats. In contrast, for $\mathcal{R}_{\text{ASCII}}$, the grid-based representation yields substantially higher performance. Accordingly, we adopt the grid format in all subsequent experiments.

\begin{table}[H]
    \centering
    \small
    \setlength{\tabcolsep}{6pt} 
    \renewcommand{\arraystretch}{1.1}
    
    \begin{tabular}{l c}
    \toprule
    \textbf{ASCII Format} & \textsc{Qwen3-30B-A3B}\\
    \midrule
    
    \multicolumn{2}{l}{\textit{ASCII $\rightarrow$ Desc}} \\
    \addlinespace[2pt]
    
    simple     & \textbf{57.1} \\
    grid    & 55.8 \\
    panel    & 54.4  \\
    
    \midrule
    
    \multicolumn{2}{l}{$\mathcal{R}_{\text{ASCII}}$} \\
    \addlinespace[2pt]
    simple     & 76.8 \\
    grid    & \textbf{83.2} \\
    panel    & 79.2 \\
    
    \bottomrule
    \end{tabular}
    
    \caption{ASCII Format Result} 
    \label{tab:ascii-format-result}
\end{table}

\section{Training configurations}
\label{appendix:training_details}
\begin{table}[H]
\centering
\small
\label{tab:training_details}
\renewcommand{\arraystretch}{1.2}

\begin{tabular}{ll}
\toprule
\textbf{Configuration} & \textbf{Value} \\
\midrule
Base Model & \textsc{Qwen3-30B-A3B} \\
Frameworks & HuggingFace Transformers, TRL, PEFT \\
Fine-tuning Method & LoRA (Rank $r=16$, $\alpha=32$) \\
Target Modules & $q, k, v, o, gate, up, down$ projections \\
Precision & bfloat16 \\
Optimizer & AdamW \\
Learning Rate & $2\times10^{-4}$ (Cosine Schedule) \\
Batch Size & 4 (per device) \\
Gradient Accumulation & 4 steps \\
Epochs & 1 \\
Dataset Split & 4,000 (Train) / 500 (Validation) \\
Hardware & 4$\times$ NVIDIA A40 \\
\bottomrule
\end{tabular}
\caption{Training configuration details.}
\end{table}

\begin{table*}[t]
\centering

\label{tab:dataset-stats}
\begin{tabular*}{0.7\textwidth}{@{\extracolsep{\fill}}lcc@{}}
\toprule
\textbf{Property} & \textbf{Full} & \textbf{Test} \\
\midrule
Instances & 20,000 & 1,000 \\
\midrule
\multicolumn{3}{@{}l}{\textit{Structural properties}} \\
\quad Components (range) & 2--8 & 2--8 \\
\quad Components ($\mu$) & 5.67 & 5.67 \\
\quad Relations (range) & 1--12 & 1--12 \\
\quad Relations ($\mu$) & 6.44 & 6.33 \\
\quad Direct relations (\%) & 52.4 & 53.2 \\
\quad Unique layouts (\%) & 41.9 & 42.4 \\
\quad Ambiguous stages ($\mu$) & 4.13 & 4.02 \\
\midrule
\multicolumn{3}{@{}l}{\textit{Query type (\%)}} \\
\quad Full & 33.3 & 33.7 \\
\quad Vertical & 33.3 & 32.6 \\
\quad Horizontal & 33.3 & 33.7 \\
\midrule
\multicolumn{3}{@{}l}{\textit{Terminology (\%)}} \\
\quad Spatial & 14.3 & 14.9 \\
\quad Cardinal & 14.3 & 14.3 \\
\quad Clock & 14.3 & 14.7 \\
\quad Spatial + Cardinal & 14.3 & 15.4 \\
\quad Spatial + Clock & 14.3 & 12.2 \\
\quad Cardinal + Clock & 14.3 & 13.5 \\
\quad All three & 14.3 & 15.0 \\
\bottomrule
\end{tabular*}
\caption{Dataset statistics for the full dataset and the evaluation subset.}
\label{Dataset_stats_in_table}
\end{table*}

\section{Confusion matrices for spatial reasoning predictions}

\begin{figure*}
    \centering
    \includegraphics[width=\linewidth]{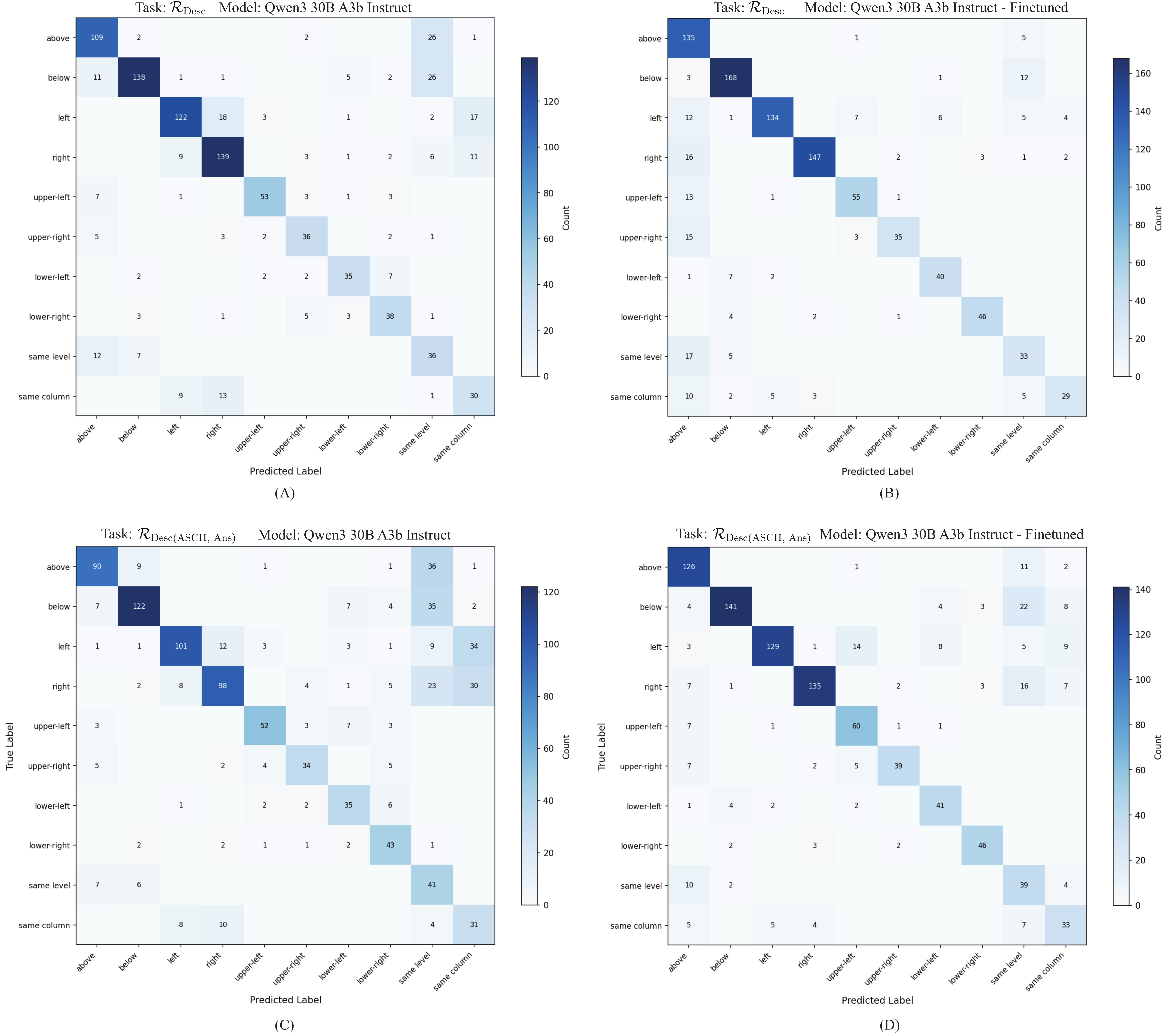}
    \caption{Confusion matrices for spatial reasoning predictions by Qwen3-30B-A3B before and after fine-tuning ({Qwen3-30B-A3B FT$_{\text{D}\rightarrow\text{A}}$}).
The top row ((A)-(B)) shows results on the text-only spatial reasoning task ($\mathcal{R}_{\text{Desc}}$), while the bottom row ((C)-(D)) shows results when ASCII is generated as an intermediate step ($\mathcal{R}_{\text{Desc:(Ans, ASCII)}}$). Left panels ((A),(C))  correspond to the base model, and right panels ((B),(D)) to the fine-tuned model. Rows indicate ground-truth spatial relations and columns indicate predicted relations. Darker diagonal cells indicate correct predictions.}
    \label{fig:confusion_matrix}
\end{figure*}

Figure \ref{fig:confusion_matrix} presents the confusion matrices for spatial reasoning predictions before and after fine-tuning on the ASCII–description conversion tasks. Fine-tuning results in stronger concentration along the diagonal and reduces systematic confusions between spatially related classes, indicating improved internal organization of spatial representations rather than uniform scaling of prediction confidence.

\section{F1 Scores}
Table~\ref{tab:model-performance-f1} reports the F1 scores for spatial reasoning tasks. In line with Table~\ref{tab:model-performance}, reasoning from ASCII alone generally outperforms reasoning from descriptions alone. While generating ASCII before answering hurts performance without fine-tuning, fine-tuning with explicit ASCII generation, either unidirectional (Desc$\rightarrow$ASCII) or bidirectional (Desc$\leftrightarrow$ASCII), consistently improves performance when only descriptions are available at inference time.
\clearpage

\begin{table*}
  \centering
  \setlength{\tabcolsep}{5pt}
  \renewcommand{\arraystretch}{1}
  \begin{tabular}{l 
                  c 
                  c 
                  c 
                  c 
                  c}
  \toprule
  \textbf{Model} 
  & \multicolumn{1}{c}{\textbf{$\mathcal{R}_{\text{Desc}}$}} 
  & \multicolumn{1}{c}{\textbf{$\mathcal{R}_{\text{ASCII}}$}} 
  & \multicolumn{1}{c}{\textbf{$\mathcal{R}_{\text{Desc(Ans, ASCII)}}$}} 
  & \multicolumn{1}{c}{\textbf{$\mathcal{R}_{\text{Desc(ASCII, Ans)}}$}} 
  & \multicolumn{1}{c}{\textbf{$\mathcal{R}_{\text{Desc+ASCII}}$}} \\
  
  \midrule
  
  \multicolumn{6}{l}{\textit{One-turn generation}} \\
  \addlinespace[2pt]
  
  \textsc{Qwen2.5-7B}        & 37.7 & 33.4 & \underline{38.6} & 33.9 & \textbf{48.3} \\
  \textsc{Llama-3-70B}       & \underline{63.3} & 62.2 & 62.3 & 52.2 & \textbf{68.8} \\
  \textsc{GPT-4.1}  & 81.2   & \underline{85.8} & 82.2     & 77.5     & \textbf{89.6} \\
  \textsc{GPT-oss-20B}       & 91.9 & 92.0 & \underline{93.2} & 90.5 & \textbf{96.8} \\
  \textsc{Qwen3-235B-A22B}   & 81.9 & \underline{86.4} & 81.5 & 79.6 & \textbf{88.0} \\
  \textsc{Qwen3-30B-A3B}     & 72.3 & \underline{80.3} & 69.6 & 66.2 & \textbf{81.3} \\
  \addlinespace[3pt]
  \textsc{Qwen3-30B-A3B FT$_{\text{D}\rightarrow\text{A}}$}       & 79.4 & 72.7 & 79.5 & \underline{76.9} & \textbf{81.9}\\
  \textsc{Qwen3-30B-A3B FT$_{\text{D}\leftarrow\text{A}}$}       & 71.4 & \underline{84.1} & 67.4 & 69.4 & \textbf{84.7}\\
  \textsc{Qwen3-30B-A3B FT$_{\text{D}\leftrightarrow\text{A}}$}       & 78.2 & 73.3 & \underline{80.0} & 79.3 & \textbf{82.6} \\
  
  \addlinespace[4pt]
  \midrule
  \addlinespace[2pt]
  
  \multicolumn{6}{l}{\textit{Two-turn generation}} \\
  \addlinespace[2pt]
  
  \textsc{Qwen2.5-7B}        &        &        & \underline{38.6} & \underline{38.6} & \textbf{46.9}\\
  \textsc{Llama-3-70B}       &        &        & \underline{61.0} & 56.2 & \textbf{67.2}\\
  \textsc{GPT-4.1}       &        &        & \underline{81.0} & 76.1 & \textbf{90.0}\\
  \textsc{GPT-oss-20B}       &        &        & \underline{92.3} & 86.8 & \textbf{94.0}\\
  \textsc{Qwen3-235B-A22B}   &        &        & \underline{83.0} & 73.1 & \textbf{85.0}\\
  \textsc{Qwen3-30B-A3B}     &        &        & \underline{71.8} & 64.1 & \textbf{78.8}\\
  \addlinespace[3pt]
  \textsc{Qwen3-30B-A3B FT$_{\text{D}\rightarrow\text{A}}$}     &        &        & \underline{76.2} & 74.9 & \textbf{78.3}\\
  \textsc{Qwen3-30B-A3B FT$_{\text{D}\leftarrow\text{A}}$}     &        &        & \underline{68.2} & 57.8 & \textbf{82.6}\\
  \textsc{Qwen3-30B-A3B FT$_{\text{D}\leftrightarrow\text{A}}$}     &        &        & \underline{77.3} & 74.0 & \textbf{80.1}\\
  \bottomrule
  \end{tabular}
  
  \caption{
  F1-score results. The best result is shown in \textbf{bold}, and the second-best is \underline{underlined}.}
  \label{tab:model-performance-f1}
\end{table*}

\clearpage
\newpage
\section{Data Generation Pipeline}
\label{appendix:data_generation_pipeline}

Pseudo-code for the spatial dataset generation pipeline with multi-dimensional balancing.
\par\vspace{1em}
\begin{algorithm}
\caption{Text2Space Dataset Generation}
\small
\label{alg:spatial_generation_balanced}
\begin{algorithmic}[1]

    \Require
    \State $N_{target}$: Target count
    \State $R_{comp}, R_{rel}$: Complexity ranges (components, relations)
\Ensure
    \State $\mathcal{D}$: Final spatial dataset

\State $\mathcal{D} \leftarrow \emptyset$
\State Initialize $Tracker$ to monitor distribution

\While{$|\mathcal{D}| < N_{target}$}
    \Statex \textbf{1. Select Parameters}
    \State Identify imbalanced categories using $Tracker$
    \State $P \leftarrow$ select parameters (e.g., node count) to balance

    \Statex \textbf{2. Build Connected Structure}
    \State $V \leftarrow \{A, B, C, \dots\}$
    \State $G \leftarrow (V, E=\emptyset)$
    \State $C \leftarrow \{ \text{Random}(V) \}$

    \While{$V \setminus C \neq \emptyset$}
        \State $u \leftarrow \text{Random}(V \setminus C)$
        \State $v \leftarrow \text{Random}(C)$
        \State $dir \leftarrow \text{Random}(\text{8 Compass Directions})$
        
        \If{edge $(u, v, dir)$ is valid}
            \State $E \leftarrow E \cup \{(u, v, dir)\}$
            \State $C \leftarrow C \cup \{u\}$
        \EndIf
    \EndWhile
    \State Add extra relations to $E$ to reach target complexity

    \Statex \textbf{3. Generate Description}
    \State $S \leftarrow \text{Random}(\{\text{Spatial, Cardinal, Clock}\})$
    \State $Desc \leftarrow \textsc{GraphToNaturalLanguage}(G, S)$

    \Statex \textbf{4. Create Query}
    \State $\mathcal{P}_{infer} \leftarrow$ Identify uniquely inferable pairs in $G$
    \State $(Q, A) \leftarrow \text{Sample pair and Type from } \mathcal{P}_{infer}$
    \State Determine if relationship is Direct vs. Inferred

    \Statex \textbf{5. Render Visualizations}
    \State $Vis \leftarrow \textsc{RenderASCII}(G, \{\text{Simple, Grid, Panel}\})$

    \Statex \textbf{6. Apply Rejection Sampling}
    \State $I \leftarrow \{G, Desc, Q, A, Vis, S\}$
    \State $\Delta \leftarrow \textsc{CalcBalanceDeviation}(I, Tracker)$
    \If{$I$ worsens balance}
        \State \textbf{Reject} with probability $P \propto \Delta^2$
    \Else
        \State $\mathcal{D} \leftarrow \mathcal{D} \cup \{I\}$
        \State Update $Tracker$ with instance $I$
    \EndIf
\EndWhile
\State \Return $\mathcal{D}$
\end{algorithmic}
\end{algorithm}

\newpage

\section{Evaluation Algorithm for ASCII and Descriptions}
Pseudo-code for evaluating generated descriptions and ASCII layouts.
\begin{algorithm}[H]
\caption{Evaluate Description vs. ASCII Layouts}
\small 
\label{alg:evaluate_desc_ascii}
\begin{algorithmic}[1]
\Require
    \State $G$: ASCII Grid (Model Output)
    \State $D$: Natural Language Description
    \State $Mode$: \text{"verify\_ascii"} or \text{"verify\_desc"}
\Ensure
    \State $R$: Result tuple (Pass, Accuracy, Breakdown)

\Statex \textbf{1. Parse Description}
\State $Segs \leftarrow \text{Split}(D, \{., ,\})$
\State $Claims \leftarrow \emptyset, \quad Bad \leftarrow \emptyset$
\For{$s \in Segs$}
    \State Match $s$ with patterns (e.g., "A is left of B")
    \If{match found}
        \State $(u, v, dir) \leftarrow \text{Extract}(s)$
        \State $Claims.add((u, v, dir))$
    \Else
        \State $Bad.add(s)$
    \EndIf
\EndFor
\State $Objs_T \leftarrow \text{UniqueObjects}(Claims)$

\Statex \textbf{2. Build Grid Graph}
\State $Pos \leftarrow \text{ScanCoordinates}(G)$
\State Normalize $Pos$ (top-left at 0,0)
\State $Truth \leftarrow \emptyset$
\For{$(u, v) \in Pos$}
    \State $\vec{d} \leftarrow Pos[u] - Pos[v]$
    \State $Truth[u, v] \leftarrow \text{VectorToDir}(\vec{d})$
\EndFor

\Statex \textbf{3. Validate Relations}
\State $Corr \leftarrow \emptyset, \quad Err \leftarrow \emptyset$
\For{$(u, v, dir) \in Claims$}
    \If{$u \notin Pos$ \textbf{or} $v \notin Pos$}
        \State $Err.add((u, v, \text{"Missing"}))$
    \Else
        \If{$dir == Truth[u, v]$}
            \State $Corr.add((u, v, dir))$
        \Else
            \State $Err.add((u, v, Truth[u, v]))$
        \EndIf
    \EndIf
\EndFor

\Statex \textbf{4. Final Scoring}
\If{$Mode == \text{"verify\_desc"}$} \Comment{ASCII is Truth}
    \State $Pass \leftarrow (Err = \emptyset)$
\Else \Comment{Text is Truth}
    \State $Extra \leftarrow Pos.keys \setminus Objs_T$
    \State $Pass \leftarrow (Err = \emptyset \land Bad = \emptyset \land Extra = \emptyset)$
\EndIf

\State $Acc \leftarrow |Corr| / |Claims|$
\State \Return $\{Pass, Acc, Corr, Err, Bad\}$
\end{algorithmic}
\end{algorithm}
\newpage

\section{Evaluation Algorithm for Consistency}
Pseudo-code for evaluating the consistency between generated ASCII layouts and query relations.
\begin{algorithm}[H]
\caption{Evaluate Consistency Between ASCII Grid and Reference relations}
\small
\label{alg:evaluate_consistency}
\begin{algorithmic}[1]
\Require
    \State $G$: ASCII representation (Model Grid)
    \State $Q$: Natural language query
    \State $L$: Reference label (relation)
\Ensure
    \State $R$: Evaluation result (correctness, objects, relations)

\Statex \textbf{1. Parse the Query}
\State Match $Q$ vs. patterns: \text{"Vertical"}, \text{"Horizontal"}, or \text{"Full"}
\State Extract target objects $X$ and $Y$ from $Q$
\If{no valid pattern found}
    \State \Return \textsc{Error}("Query parsing failed")
\EndIf

\Statex \textbf{2. Parse the ASCII Grid}
\State Scan $G$ to find coordinates for all characters
\State $Pos \leftarrow \{(obj, x, y) \mid obj \text{ found in } G\}$
\If{$X \notin Pos$ \textbf{or} $Y \notin Pos$}
    \State \Return \textsc{Error}("Queried object missing in grid")
\EndIf
\State Normalize $Pos$ so top-left object is at $(0,0)$

\Statex \textbf{3. Infer Relation from Grid}
\State $(x_1, y_1) \leftarrow Pos[X], \quad (x_2, y_2) \leftarrow Pos[Y]$
\State $Rel_{inf} \leftarrow \emptyset$

\If{Type is \text{"Vertical"}}
    \If{$y_1 < y_2$} $Rel_{inf} \leftarrow \text{"above"}$
    \ElsIf{$y_1 > y_2$} $Rel_{inf} \leftarrow \text{"below"}$
    \Else \ $Rel_{inf} \leftarrow \text{"same level"}$
    \EndIf
\ElsIf{Type is \text{"Horizontal"}}
    \If{$x_1 < x_2$} $Rel_{inf} \leftarrow \text{"left"}$
    \ElsIf{$x_1 > x_2$} $Rel_{inf} \leftarrow \text{"right"}$
    \Else \ $Rel_{inf} \leftarrow \text{"same column"}$
    \EndIf
\Else \Comment{Full Spatial Query}
    \State Compute vector $\vec{v} = (x_1 - x_2, y_1 - y_2)$
    \State $Rel_{inf} \leftarrow$ Map $\vec{v}$ to \{upper-left, below, right, ...\}
\EndIf

\Statex \textbf{4. Compare with Reference Label}
\State $Match \leftarrow (\text{LowerCase}(Rel_{inf}) == \text{LowerCase}(L))$
\State $R \leftarrow \{ \text{Pass}: Match, \text{Exp}: L, \text{Act}: Rel_{inf} \}$
\State \Return $R$
\end{algorithmic}
\end{algorithm}

\clearpage
\newpage
\section{Human Validation}
\label{appendix:human validation}
\paragraph{Human Validation Setup.} To quantify the impact of different input formats on spatial reasoning accuracy, we conducted a controlled human validation study with 11 evaluators divided into two groups: Team A ($n=7$) and Team B ($n=4$). We curated three distinct problem sets (Set X, Set Y, and Set Z), each containing 10 spatially complex examples. The design details are shown in Table~\ref{tab:human_study_design}

The study utilized a counterbalanced within-subjects design to disentangle the effect of input format from question difficulty. All participants completed three sessions:

\begin{enumerate}
    \item \textbf{Session 2 (Visual Baseline):} Both teams evaluated Set Z using the \textbf{ASCII-only} format. Performance was highly consistent across groups (Team A: 0.986, Team B: 0.950), confirming that both teams possessed comparable visual reasoning baselines.
    \item \textbf{Sessions 1 \& 3 (Crossover):} We swapped the assignment of problem sets and formats between teams. Team A solved Set X using \textbf{Description-only} (Session 1) and Set Y using \textbf{Description+ASCII} (Session 3). Conversely, Team B solved Set Y using \textbf{Description-only} (Session 1) and Set X using \textbf{Description+ASCII} (Session 3).
\end{enumerate}

This cross-over design ensures that Sets X and Y were evaluated in both text-only and combined formats, allowing for a direct comparison of performance on semantically identical questions.

\begin{table}[h]
    \centering
    \small
    \setlength{\tabcolsep}{3pt} 
    \renewcommand{\arraystretch}{1.2}
    
    \begin{tabular}{@{} l p{2.5cm} p{2.5cm} @{}}
    \toprule
    \textbf{Task Sessions} & \textbf{Team A} ($n=7$) & \textbf{Team B} ($n=4$) \\
    \midrule
    
    \textbf{Session 1} & \textbf{Description Only} & \textbf{Description Only} \\
    \textit{\scriptsize (Single Format)} & \small (Set X) & \small (Set Y) \\
    \addlinespace[4pt]
    
    \textbf{Session 2} & \textbf{ASCII Only} & \textbf{ASCII Only} \\
    \textit{\scriptsize (Visual Baseline)} & \small (Set Z) & \small (Set Z) \\
    \addlinespace[4pt]
    
    \textbf{Session 3} & \textbf{Desc + ASCII} & \textbf{Desc + ASCII} \\
    \textit{\scriptsize (Combined)} & \small (Set Y) & \small (Set X) \\
    
    \bottomrule
    \end{tabular}
    
    \caption{Human Validation Study Design. The experiment followed a three-session protocol with counterbalanced question sets.} 
    \label{tab:human_study_design}
\end{table}

\newpage
\paragraph{Validation Results}
Table~\ref{tab:human validation results} shows the substantial improvement achieved when reasoning from an ASCII-augmented description compared to reasoning from the description alone.

\begin{table}[h]
    \centering
    \small
    \setlength{\tabcolsep}{1pt} 
    \renewcommand{\arraystretch}{1.2}
    \begin{tabular}{@{} l c c c @{}}
    \toprule
    \textbf{Question Set} & \textbf{Desc Only} & \textbf{Desc + ASCII} & \textbf{Improvement} \\
     & (NL Format) & (Combined) & \\
    \midrule
    Set X & 0.857 \footnotesize{(Team A)} & 0.975 \footnotesize{(Team B)} & +11.8\% \\
    Set Y & 0.875 \footnotesize{(Team B)} & 0.986 \footnotesize{(Team A)} & +11.1\% \\
    \midrule
    \textbf{Average} & \textbf{0.866} & \textbf{0.981} & \textbf{+11.5\%} \\
    \bottomrule
    \end{tabular}
    
    \caption{Human performance comparison across different input formats. By counterbalancing Sets X and Y, we isolate the improvement attributed to the format.}
    \label{tab:human validation results}
\end{table}


\begin{figure*}
    \centering
    \includegraphics[width=\linewidth]{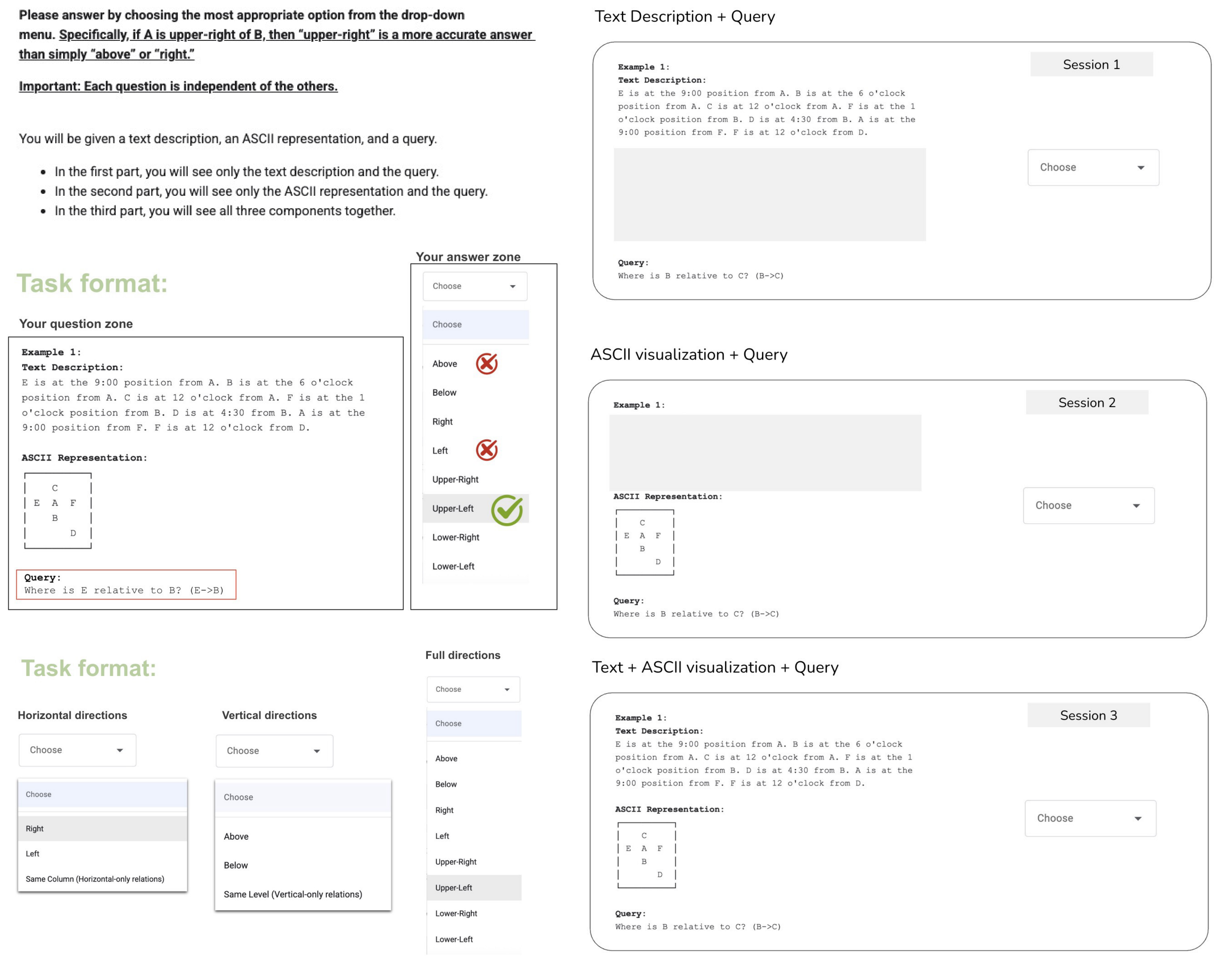}
    \caption{Overview of the human validation interface and instruction protocol. Participants solved spatial reasoning tasks across three counterbalanced sessions: (1) Text Description Only, (2) ASCII Visualization Only, and (3) Combined Text + ASCII. We use Google Form to collect the response from the participants. The interface requires participants to select the most precise directional term from a standardized drop-down menu.}
    \label{fig:placeholder}
\end{figure*}

\clearpage

\section{Prompts and Outputs}
\subsection{Task Description $\rightarrow$ ASCII}
\begin{lstlisting}
---------
SYSTEM
---------
Assume you have expertise in spatial reasoning. Your task is to answer the following question to the best of your knowledge.
You will receive a scene description.
Your job is to generate the requested ASCII visualization and return the result as a compact JSON object.

Description -> ASCII Diagram.
Read the textual scene description and render the requested ASCII layout.

- Treat the canvas like a map: the first printed row is the top, and the last row is the bottom.
- The first column is the left edge; the last column is the right edge.
- Above = higher row; below = lower row (do not invert the vertical axis).
- Left = smaller column index; right = larger column index (do not mirror horizontally).

Formats:
- grid: <grid using + - | for borders>

Rendering & Positioning Rules

Use given uppercase letters (A, B, C, ...); don't invent new ones.
Grid size is flexible - expand rows/cols as needed to fit all points.
Place all letters; if two overlap, stack them (e.g., "AF").

Relations are first -> second (e.g., "D to F" means D relative to F).

Positioning:
above/below -> same column
left/right -> same level
diagonals (upper/lower-left/right) -> one cell diagonally.
Keep perfect alignment for cardinal directions.
Adjacent items occupy neighboring cells; no diagonal drift.
Maintain uniform cell sizes and consistent borders.

Keep row alignment when depicting left/right relations; do not shift labels vertically.
Keep column alignment when depicting above/below relations; do not shift labels horizontally.
For diagonal relations, ensure the two entities occupy different rows and different columns.
Avoid inconsistent cell widths and do not introduce extra or missing letters.

Additional ASCII Guidance:
- Label every entity exactly as named in the description.

Input Notes:
- Use only the provided description to infer entity positions.
- Do not invent additional entities or relations when information is missing.

Output Notes:
- Produce ASCII views that faithfully match the described scene.

OUTPUT SCHEMA (JSON):
{
  "ascii": {
    "grid": "<grid using + - | for borders>"
  }
}

Return ONLY a compact JSON object.
Do not include markdown fences or any extra text.

---------
USER
---------
Description:
B is to the lower-right of A. D is to the right of B. C is at the bottom-right of A. C is located below B.

---------
ASSISTANT
---------
{"ascii":{"grid":"+--+--+--+\n|A |  |  |\n+--+--+--+\n|  |B |D |\n+--+--+--+\n|  |C |  |\n+--+--+--+"}}
\end{lstlisting}

\subsection{Task ASCII $\rightarrow$ Description}
\begin{lstlisting}
---------
SYSTEM
---------
Assume you have expertise in spatial reasoning. Your task is to answer the following question to the best of your knowledge.
You will receive ASCII diagram(s).
Your job is to write a concise spatial summary and return the result as a compact JSON object.

ASCII Diagram -> Description.
Interpret the ASCII evidence and rewrite it as a concise natural-language description.

- Treat the canvas like a map: the first printed row is the top, and the last row is the bottom.
- The first column is the left edge; the last column is the right edge.
- Above = higher row; below = lower row (do not invert the vertical axis).
- Left = smaller column index; right = larger column index (do not mirror horizontally).

Formats:
- grid: <grid using + - | for borders>

Input Notes:
- Treat the ASCII diagram as authoritative evidence.

Output Notes:
- Summarize the spatial arrangement in clear declarative sentences.

Description Guidelines:
- Mention every entity shown in the ASCII input and describe their relative placement.
- Keep the description short and factual; avoid speculative language.
- The spatial summary will consist of sentences in the format "A is [relation] B." Each sentence should describe the relationship between exactly two objects.

OUTPUT SCHEMA (JSON):
{
  "description": "<natural language spatial summary>"
}

Return ONLY a compact JSON object.
Do not include markdown fences or any extra text.

---------
USER
---------
ASCII Reference:
grid:
+--+--+--+
|A |  |  |
+--+--+--+
|  |B |D |
+--+--+--+
|  |C |  |
+--+--+--+

---------
ASSISTANT
---------
{"description":"B is to the lower-right of A. D is to the right of B. C is at the bottom-right of A. C is located below B."}
\end{lstlisting}

\subsection{Task $\mathcal{R}_{\text{Desc}}$}
\begin{lstlisting}
---------
SYSTEM
---------
Assume you have expertise in spatial reasoning. Your task is to answer the following question to the best of your knowledge.
You will receive a directional question and a scene description.
Your job is to produce the correct directional answer and return the result as a compact JSON object.

Note that the possible directions are defined as {left, right, above, below, upper-left, upper-right, lower-left, lower-right, same column, same level}, and there will be only one correct answer.

Description + Query -> Answer.
Answer the spatial relationship question using the textual description.

10 Distinct (Mutually Exclusive) Directions
Pure Directions - perfectly aligned:
- left: directly left, no vertical offset
- right: directly right, no vertical offset
- above: directly above, no horizontal offset
- below: directly below, no vertical offset
Diagonal Directions - both vertical and horizontal offset: upper-left, upper-right, lower-left, lower-right
Horizontal Alignment ONLY: same level; Vertical Alignment ONLY: same column

Input Notes:
- Use the description to reason about the positions of the queried entities.

Output Notes:
- Return the spatial relation as a single direction string.

Direction Notes:
- Ensure the answer directly references the entities named in the question.

OUTPUT SCHEMA (JSON):
{
  "answer": "<direction>"
}

Return ONLY a compact JSON object.
Do not include markdown fences or any extra text.

---------
USER
---------
Description:
B is to the lower-right of A. D is to the right of B. C is at the bottom-right of A. C is located below B.

Question: Where is C horizontally relative to D?

---------
ASSISTANT
---------
{"answer":"left"}  
\end{lstlisting}

\subsection{Task $\mathcal{R}_{\text{ASCII}}$}
\begin{lstlisting}
---------
SYSTEM
---------
Assume you have expertise in spatial reasoning. Your task is to answer the following question to the best of your knowledge.
You will receive a directional question and ASCII diagram(s).
Your job is to produce the correct directional answer and return the result as a compact JSON object.

Note that the possible directions are defined as {left, right, above, below, upper-left, upper-right, lower-left, lower-right, same column, same level}, and there will be only one correct answer.

ASCII + Query -> Answer.
Answer the spatial relationship question using the ASCII diagram as the primary evidence.

10 Distinct (Mutually Exclusive) Directions
Pure Directions - perfectly aligned:
- left: directly left, no vertical offset
- right: directly right, no vertical offset
- above: directly above, no horizontal offset
- below: directly below, no vertical offset
Diagonal Directions - both vertical and horizontal offset: upper-left, upper-right, lower-left, lower-right
Horizontal Alignment ONLY: same level; Vertical Alignment ONLY: same column

- Treat the canvas like a map: the first printed row is the top, and the last row is the bottom.
- The first column is the left edge; the last column is the right edge.
- Above = higher row; below = lower row (do not invert the vertical axis).
- Left = smaller column index; right = larger column index (do not mirror horizontally).

Formats:
- grid: <grid using + - | for borders>

Input Notes:
- Use the ASCII diagram as ground truth; the query specifies the entities to compare.

Output Notes:
- Provide a single directional answer from the defined set.

Direction Notes:
- Base the answer on the ASCII layout; ignore assumptions that contradict the diagram.

OUTPUT SCHEMA (JSON):
{
  "answer": "<direction>"
}

Return ONLY a compact JSON object.
Do not include markdown fences or any extra text.

---------
USER
---------
ASCII Reference:
grid:
+--+--+--+
|A |  |  |
+--+--+--+
|  |B |D |
+--+--+--+
|  |C |  |
+--+--+--+

Question: Where is C horizontally relative to D?

---------
ASSISTANT
---------
{"answer":"left"}
\end{lstlisting}

\subsection{Task $\mathcal{R}_{\text{Desc(Ans, ASCII)}}$}
\textit{One-turn Generation}
\begin{lstlisting}
---------
SYSTEM
---------
Assume you have expertise in spatial reasoning. Your task is to answer the following question to the best of your knowledge.
You will receive a directional question and a scene description.
Your job is to produce the correct directional answer and generate the requested ASCII visualization and return the result as a compact JSON object.

Note that the possible directions are defined as {left, right, above, below, upper-left, upper-right, lower-left, lower-right, same column, same level}, and there will be only one correct answer.

Description + Query -> ASCII + Answer.
Answer the spatial question and draw ASCII that matches the same layout.

10 Distinct (Mutually Exclusive) Directions
Pure Directions - perfectly aligned:
- left: directly left, no vertical offset
- right: directly right, no vertical offset
- above: directly above, no horizontal offset
- below: directly below, no vertical offset
Diagonal Directions - both vertical and horizontal offset: upper-left, upper-right, lower-left, lower-right
Horizontal Alignment ONLY: same level; Vertical Alignment ONLY: same column

- Treat the canvas like a map: the first printed row is the top, and the last row is the bottom.
- The first column is the left edge; the last column is the right edge.
- Above = higher row; below = lower row (do not invert the vertical axis).
- Left = smaller column index; right = larger column index (do not mirror horizontally).

Formats:
- grid: <grid using + - | for borders>

Rendering & Positioning Rules

Use given uppercase letters (A, B, C, ...); don't invent new ones.
Grid size is flexible - expand rows/cols as needed to fit all points.
Place all letters; if two overlap, stack them (e.g., "AF").

Relations are first -> second (e.g., "D to F" means D relative to F).

Positioning:
above/below -> same column
left/right -> same level
diagonals (upper/lower-left/right) -> one cell diagonally.
Keep perfect alignment for cardinal directions.
Adjacent items occupy neighboring cells; no diagonal drift.
Maintain uniform cell sizes and consistent borders.

Keep row alignment when depicting left/right relations; do not shift labels vertically.
Keep column alignment when depicting above/below relations; do not shift labels horizontally.
For diagonal relations, ensure the two entities occupy different rows and different columns.
Avoid inconsistent cell widths and do not introduce extra or missing letters.

Additional ASCII Guidance:
- Ensure the ASCII visualization aligns with the answer you provide.

Determine the correct answer first, then draw ASCII that matches it.

Input Notes:
- Use the description to understand the scene before answering.
- Ensure the ASCII rendering is consistent with the textual evidence.

Output Notes:
- Provide both the answer and the requested ASCII views in the JSON output.

Direction Notes:
- Check the ASCII you generate so it supports the chosen answer.

OUTPUT SCHEMA (JSON):
{
  "answer": "<direction>",
  "ascii": {
    "grid": "<grid using + - | for borders>"
  }
}

Return ONLY a compact JSON object.
Do not include markdown fences or any extra text.

---------
USER
---------
Description:
B is to the lower-right of A. D is to the right of B. C is at the bottom-right of A. C is located below B.

Question: Where is C horizontally relative to D?

---------
ASSISTANT
---------
{"answer":"left","ascii":{"grid":"+--+--+--+\n|A |  |  |\n+--+--+--+\n|  |B |D |\n+--+--+--+\n|  |C |  |\n+--+--+--+"}}
\end{lstlisting}
\textit{Two-turn Generation}
\begin{lstlisting}
---------
SYSTEM
---------
Assume you have expertise in spatial reasoning. Your task is to answer the following question to the best of your knowledge.
You will receive a directional question and a scene description.
Your task is to provide both the correct directional answer and the corresponding ASCII visualization, including both in a single compact JSON object.

This is a two-turn task: Answer First, Then Generate ASCII.

- Turn 1: Answer the spatial question based on reasoning.
- Turn 2: Generate an ASCII visualization that matches your answer.

## Turn 1: Answer the Question

The possible directions are {left, right, above, below, upper-left, upper-right, lower-left, lower-right, same column, same level}. There is only one correct answer.

10 Distinct (Mutually Exclusive) Directions
Pure Directions - perfectly aligned:
- left: directly left, no vertical offset
- right: directly right, no vertical offset
- above: directly above, no horizontal offset
- below: directly below, no vertical offset
Diagonal Directions - both vertical and horizontal offset: upper-left, upper-right, lower-left, lower-right
Horizontal Alignment ONLY: same level; Vertical Alignment ONLY: same column

## Turn 2: Generate ASCII Visualization

Formats:
- grid: <grid using + - | for borders>

Rendering & Positioning Rules

Use given uppercase letters (A, B, C, ...); don't invent new ones.
Grid size is flexible - expand rows/cols as needed to fit all points.
Place all letters; if two overlap, stack them (e.g., "AF").

Relations are first -> second (e.g., "D to F" means D relative to F).

Positioning:
above/below -> same column
left/right -> same level
diagonals (upper/lower-left/right) -> one cell diagonally.
Keep perfect alignment for cardinal directions.
Adjacent items occupy neighboring cells; no diagonal drift.
Maintain uniform cell sizes and consistent borders.

Keep row alignment when depicting left/right relations; do not shift labels vertically.
Keep column alignment when depicting above/below relations; do not shift labels horizontally.
For diagonal relations, ensure the two entities occupy different rows and different columns.
Avoid inconsistent cell widths and do not introduce extra or missing letters.

- Treat the canvas like a map: the first printed row is the top, and the last row is the bottom.
- The first column is the left edge; the last column is the right edge.
- Above = higher row; below = lower row (do not invert the vertical axis).
- Left = smaller column index; right = larger column index (do not mirror horizontally).

Output Notes:
- Determine the answer based on spatial reasoning from the description.
- Then create ASCII that accurately reflects your answer.
- Turn 1: Provide the directional answer as a JSON object with "answer" key.
- Turn 2: Provide the ASCII views as a JSON object with "ascii" key.

OUTPUT SCHEMA (JSON):
{
  "answer": "<direction>",
  "ascii": {
    "grid": "<grid using + - | for borders>"
  }
}

Return ONLY a compact JSON object.
Do not include markdown fences or any extra text.

---------
USER
---------
Description:
B is to the lower-right of A. D is to the right of B. C is at the bottom-right of A. C is located below B.

Question: Where is C horizontally relative to D?

Please answer this question.

---------
ASSISTANT
---------
{"answer":"left"}

---------
USER
---------
Now, generate an ASCII visualization that matches your answer and shows this spatial arrangement.

---------
ASSISTANT
---------
{"ascii":{"grid":"+--+--+--+\n|A |  |  |\n+--+--+--+\n|  |B |D |\n+--+--+--+\n|  |C |  |\n+--+--+--+"}}
\end{lstlisting}
\subsection{Task $\mathcal{R}_{\text{Desc(ASCII, Ans)}}$}
\textit{One-turn Generation}
\begin{lstlisting}
---------
SYSTEM
---------
Assume you have expertise in spatial reasoning. Your task is to answer the following question to the best of your knowledge.
You will receive a directional question and a scene description.
Your job is to produce the correct directional answer and generate the requested ASCII visualization and return the result as a compact JSON object.

Note that the possible directions are defined as {left, right, above, below, upper-left, upper-right, lower-left, lower-right, same column, same level}, and there will be only one correct answer.

Description + Query -> ASCII + Answer.
Answer the spatial question and draw ASCII that matches the same layout.

10 Distinct (Mutually Exclusive) Directions
Pure Directions - perfectly aligned:
- left: directly left, no vertical offset
- right: directly right, no vertical offset
- above: directly above, no horizontal offset
- below: directly below, no vertical offset
Diagonal Directions - both vertical and horizontal offset: upper-left, upper-right, lower-left, lower-right
Horizontal Alignment ONLY: same level; Vertical Alignment ONLY: same column

- Treat the canvas like a map: the first printed row is the top, and the last row is the bottom.
- The first column is the left edge; the last column is the right edge.
- Above = higher row; below = lower row (do not invert the vertical axis).
- Left = smaller column index; right = larger column index (do not mirror horizontally).

Formats:
- grid: <grid using + - | for borders>

Rendering & Positioning Rules

Use given uppercase letters (A, B, C, ...); don't invent new ones.
Grid size is flexible - expand rows/cols as needed to fit all points.
Place all letters; if two overlap, stack them (e.g., "AF").

Relations are first -> second (e.g., "D to F" means D relative to F).

Positioning:
above/below -> same column
left/right -> same level
diagonals (upper/lower-left/right) -> one cell diagonally.
Keep perfect alignment for cardinal directions.
Adjacent items occupy neighboring cells; no diagonal drift.
Maintain uniform cell sizes and consistent borders.

Keep row alignment when depicting left/right relations; do not shift labels vertically.
Keep column alignment when depicting above/below relations; do not shift labels horizontally.
For diagonal relations, ensure the two entities occupy different rows and different columns.
Avoid inconsistent cell widths and do not introduce extra or missing letters.

Additional ASCII Guidance:
- Ensure the ASCII visualization aligns with the answer you provide.

Generate the ASCII visualization before finalizing the answer.

Input Notes:
- Use the description to understand the scene before answering.
- Ensure the ASCII rendering is consistent with the textual evidence.

Output Notes:
- Provide both the answer and the requested ASCII views in the JSON output.

Direction Notes:
- Check the ASCII you generate so it supports the chosen answer.

OUTPUT SCHEMA (JSON):
{
  "ascii": {
    "grid": "<grid using + - | for borders>"
  },
  "answer": "<direction>"
}

Return ONLY a compact JSON object.
Do not include markdown fences or any extra text.

---------
USER
---------
Description:
B is to the lower-right of A. D is to the right of B. C is at the bottom-right of A. C is located below B.

Question: Where is C horizontally relative to D?

---------
ASSISTANT
---------
{"ascii":{"grid":"+--+--+--+\n|A |  |  |\n+--+--+--+\n|  |B |D |\n+--+--+--+\n|  |C |  |\n+--+--+--+"},"answer":"left"}
\end{lstlisting}

\textit{Two-turn Generation}
\begin{lstlisting}
---------
SYSTEM
---------
Assume you have expertise in spatial reasoning. Your task is to answer the following question to the best of your knowledge.
You will receive a directional question and a scene description.
Your task is to provide both the correct directional answer and the corresponding ASCII visualization, including both in a single compact JSON object.

This is a two-turn task: Generate ASCII First, Then Answer.

- Turn 1: Generate an ASCII visualization from the scene description.
- Turn 2: Answer the spatial question using the ASCII you generated as a reference.

## Turn 1: Generate ASCII Visualization

Formats:
- grid: <grid using + - | for borders>

Rendering & Positioning Rules

Use given uppercase letters (A, B, C, ...); don't invent new ones.
Grid size is flexible - expand rows/cols as needed to fit all points.
Place all letters; if two overlap, stack them (e.g., "AF").

Relations are first -> second (e.g., "D to F" means D relative to F).

Positioning:
above/below -> same column
left/right -> same level
diagonals (upper/lower-left/right) -> one cell diagonally.
Keep perfect alignment for cardinal directions.
Adjacent items occupy neighboring cells; no diagonal drift.
Maintain uniform cell sizes and consistent borders.

Keep row alignment when depicting left/right relations; do not shift labels vertically.
Keep column alignment when depicting above/below relations; do not shift labels horizontally.
For diagonal relations, ensure the two entities occupy different rows and different columns.
Avoid inconsistent cell widths and do not introduce extra or missing letters.

- Treat the canvas like a map: the first printed row is the top, and the last row is the bottom.
- The first column is the left edge; the last column is the right edge.
- Above = higher row; below = lower row (do not invert the vertical axis).
- Left = smaller column index; right = larger column index (do not mirror horizontally).

## Turn 2: Answer the Question

The possible directions are {left, right, above, below, upper-left, upper-right, lower-left, lower-right, same column, same level}. There is only one correct answer.

10 Distinct (Mutually Exclusive) Directions
Pure Directions - perfectly aligned:
- left: directly left, no vertical offset
- right: directly right, no vertical offset
- above: directly above, no horizontal offset
- below: directly below, no vertical offset
Diagonal Directions - both vertical and horizontal offset: upper-left, upper-right, lower-left, lower-right
Horizontal Alignment ONLY: same level; Vertical Alignment ONLY: same column

Output Notes:
- Generate the ASCII visualization before attempting to answer.
- Use the ASCII you generated to determine the correct answer.
- Turn 1: Provide the ASCII views as a JSON object with "ascii" key.
- Turn 2: Provide the directional answer as a JSON object with "answer" key.

OUTPUT SCHEMA (JSON):
{
  "answer": "<direction>",
  "ascii": {
    "grid": "<grid using + - | for borders>"
  }
}

Return ONLY a compact JSON object.
Do not include markdown fences or any extra text.

---------
USER
---------
Description:
B is to the lower-right of A. D is to the right of B. C is at the bottom-right of A. C is located below B.

Question: Where is C horizontally relative to D?

Please generate an ASCII visualization of this scene.

---------
ASSISTANT
---------
{"ascii":{"grid":"+--+--+--+\n|A |  |  |\n+--+--+--+\n|  |B |D |\n+--+--+--+\n|  |C |  |\n+--+--+--+"}}

---------
USER
---------
Now, based on the ASCII visualization you generated, answer the question: Where is C horizontally relative to D?

---------
ASSISTANT
---------
{"answer":"left"}
\end{lstlisting}

\subsection{Task $\mathcal{R}_{\text{Desc:(ASCII, Ans)}}$}
\textit{One-turn Generation}
\begin{lstlisting}
---------
SYSTEM
---------
Assume you have expertise in spatial reasoning. Your task is to answer the following question to the best of your knowledge.
You will receive a directional question, a scene description, and ASCII diagram(s).
Your job is to produce the correct directional answer and return the result as a compact JSON object.

Note that the possible directions are defined as {left, right, above, below, upper-left, upper-right, lower-left, lower-right, same column, same level}, and there will be only one correct answer.

Description + ASCII + Query -> Answer.
Combine textual and ASCII evidence to answer the spatial question.

10 Distinct (Mutually Exclusive) Directions
Pure Directions - perfectly aligned:
- left: directly left, no vertical offset
- right: directly right, no vertical offset
- above: directly above, no horizontal offset
- below: directly below, no vertical offset
Diagonal Directions - both vertical and horizontal offset: upper-left, upper-right, lower-left, lower-right
Horizontal Alignment ONLY: same level; Vertical Alignment ONLY: same column

- Treat the canvas like a map: the first printed row is the top, and the last row is the bottom.
- The first column is the left edge; the last column is the right edge.
- Above = higher row; below = lower row (do not invert the vertical axis).
- Left = smaller column index; right = larger column index (do not mirror horizontally).

Formats:
- grid: <grid using + - | for borders>

Input Notes:
- Cross-check the description and ASCII diagram; treat ASCII as authoritative when conflicts arise.

Output Notes:
- Output a single canonical direction that matches the combined evidence.

Direction Notes:
- If the two evidence sources disagree, follow the ASCII diagram.

OUTPUT SCHEMA (JSON):
{
  "answer": "<direction>"
}

Return ONLY a compact JSON object.
Do not include markdown fences or any extra text.

---------
USER
---------
Description:
B is to the lower-right of A. D is to the right of B. C is at the bottom-right of A. C is located below B.

ASCII Reference:
grid:
+--+--+--+
|A |  |  |
+--+--+--+
|  |B |D |
+--+--+--+
|  |C |  |
+--+--+--+

Question: Where is C horizontally relative to D?

---------
ASSISTANT
---------
{"answer":"left"}
\end{lstlisting}

\textit{Two-turn Generation} with ground truth ASCII
\begin{lstlisting}
---------
SYSTEM
---------
Assume you have expertise in spatial reasoning. Your task is to answer the following question to the best of your knowledge.
You will receive a directional question and a scene description.
Your task is to provide both the correct directional answer and the corresponding ASCII visualization, including both in a single compact JSON object.

This is a two-turn task: Generate ASCII First, Then Answer.

- Turn 1: Generate an ASCII visualization from the scene description.
- Turn 2: Answer the spatial question using the ASCII you generated as a reference.

## Turn 1: Generate ASCII Visualization

Formats:
- grid: <grid using + - | for borders>

Rendering & Positioning Rules

Use given uppercase letters (A, B, C, ...); don't invent new ones.
Grid size is flexible - expand rows/cols as needed to fit all points.
Place all letters; if two overlap, stack them (e.g., "AF").

Relations are first -> second (e.g., "D to F" means D relative to F).

Positioning:
above/below -> same column
left/right -> same level
diagonals (upper/lower-left/right) -> one cell diagonally.
Keep perfect alignment for cardinal directions.
Adjacent items occupy neighboring cells; no diagonal drift.
Maintain uniform cell sizes and consistent borders.

Keep row alignment when depicting left/right relations; do not shift labels vertically.
Keep column alignment when depicting above/below relations; do not shift labels horizontally.
For diagonal relations, ensure the two entities occupy different rows and different columns.
Avoid inconsistent cell widths and do not introduce extra or missing letters.

- Treat the canvas like a map: the first printed row is the top, and the last row is the bottom.
- The first column is the left edge; the last column is the right edge.
- Above = higher row; below = lower row (do not invert the vertical axis).
- Left = smaller column index; right = larger column index (do not mirror horizontally).

## Turn 2: Answer the Question

The possible directions are {left, right, above, below, upper-left, upper-right, lower-left, lower-right, same column, same level}. There is only one correct answer.

10 Distinct (Mutually Exclusive) Directions
Pure Directions - perfectly aligned:
- left: directly left, no vertical offset
- right: directly right, no vertical offset
- above: directly above, no horizontal offset
- below: directly below, no vertical offset
Diagonal Directions - both vertical and horizontal offset: upper-left, upper-right, lower-left, lower-right
Horizontal Alignment ONLY: same level; Vertical Alignment ONLY: same column

Output Notes:
- Generate the ASCII visualization before attempting to answer.
- Use the ASCII you generated to determine the correct answer.
- Turn 1: Provide the ASCII views as a JSON object with "ascii" key.
- Turn 2: Provide the directional answer as a JSON object with "answer" key.

OUTPUT SCHEMA (JSON):
{
  "answer": "<direction>",
  "ascii": {
    "grid": "<grid using + - | for borders>"
  }
}

Return ONLY a compact JSON object.
Do not include markdown fences or any extra text.

---------
USER
---------
Description:
B is to the lower-right of A. D is to the right of B. C is at the bottom-right of A. C is located below B.

Question: Where is C horizontally relative to D?

Please generate an ASCII visualization of this scene.

---------
ASSISTANT
---------
{"ascii":{"grid":"+--+--+--+\n|A |  |  |\n+--+--+--+\n|  |B |D |\n+--+--+--+\n|  |C |  |\n+--+--+--+"}}

---------
USER
---------
Now, based on the ASCII visualization you generated, answer the question: Where is C horizontally relative to D?

---------
ASSISTANT
---------
{"answer":"left"}
\end{lstlisting}

\newpage

\end{document}